# FEM-Informed Hypergraph Neural Networks for Efficient Elastoplasticity

Jianchuan Yang[a], Xi Chen[b], Jidong Zhao[a]†

[a]*Department of Civil and Environmental Engineering, Hong Kong University of Science and Technology, Clear Water Bay, Kowloon, Hong Kong, China*

[b]*Department of Mechanical and Aerospace Engineering, The Hong Kong University of Science and Technology, Clear Water Bay, Kowloon, Hong Kong, China*

†Corresponding author. Email: jzhao@ust.hk

## Abstract：

Graph neural networks (GNNs) naturally align with sparse operators and unstructured discretizations, making them a promising paradigm for physics-informed machine learning in computational mechanics. Motivated by discrete physics losses and Hierarchical Deep Learning Neural Network (HiDeNN) constructions, we embed finite-element (FEM) computations at nodes and Gauss points directly into message-passing layers and propose a numerically consistent FEM-Informed Hypergraph Neural Networks (FHGNN). Similar to conventional physics-informed neural networks (PINNs), training is purely physics-driven and requires no labeled data: the input is a node–element hypergraph whose edges encode mesh connectivity. Guided by empirical results and condition-number analysis, we adopt an efficient variational loss. Validated on 3D benchmarks, including cyclic loading with isotropic/kinematic hardening, the proposed method delivers substantially improved accuracy and efficiency over recent, competitive PINN variants. By leveraging GPU-parallel tensor operations and the discrete representation, it scales effectively to large elastoplastic problems and can be competitive with, or faster than, multi-core FEM implementations at comparable accuracy. This work establishes a foundation for scalable, physics-embedded learning in nonlinear solid mechanics.

# 1 Introduction

Machine learning has achieved notable success in data-driven domains such as computer vision and natural language processing. In contrast, solving partial differential equations (PDEs) in engineering applications critically relies on the incorporation of physical priors, which constitutes a central theme of scientific machine learning. Physics-Informed Neural Networks (PINNs) [1,2] are a representative approach that enforces governing equations by penalizing PDE residuals in the loss function, enabling a wide range of forward and inverse problems, including fluid mechanics [3–6], solid mechanics [7,8], and materials science [9,10]. In solid mechanics, early PINN studies primarily employed automatic differentiation (AD) to construct physics-based loss functions and optimize multilayer perceptrons (MLPs), demonstrating feasibility in linear elasticity [11], hyperelasticity [12], and elastoplasticity [13,14]. However, for nonlinear elastoplasticity, challenges remain in prediction accuracy and computational cost. Recent efforts have therefore focused on more expressive architectures and the integration of numerical algorithms to enhance performance.

For MLPs, the dense connections and the large computational graphs induced by AD, often lead to high training costs and slow convergence. Enforcing boundary conditions is also nontrivial: gradient imbalance between boundary penalties and physics residuals can hinder optimization. Convolutional neural networks (CNNs) can alleviate part of these issues by adopting discrete representations with sparse connectivity and weight sharing, enabling efficient parallel evaluation. Moreover, numerical discretization schemes (e.g., finite differences [15] and finite volumes [5,16]) can be used to construct losses without high-order AD, thereby reducing computational overhead. Nonetheless, CNN-based approaches typically rely on structured grids and regular domains, which limits their applicability to complex geometries and hampers local refinement required to capture stress concentrations or plastic zones. Coordinate-transformation-based strategies, such as PhyGeoNet [17] and JacobiNet [18], map irregular domains to regular ones, but to date have been validated primarily for relatively simple geometries.

By representing mesh nodes and their connectivity as graph-structured data, graph neural networks (GNNs) are naturally compatible with complex geometries and unstructured meshes, while retaining the benefits of discrete operators and parallel computation. Since their inception, a variety of graph convolutional kernels have been proposed [19–21], which has stimulated increasing applications in computational mechanics [22,23]. He [24] demonstrated through numerical experiments on linear elasticity and Neo-Hookean materials that GCN-based formulations can achieve improved performance in the deep energy method (DEM). Gao [25] employed Chebyshev graph convolutions to learn a nonlinear mapping from coordinates to displacements, proposing a discrete PINN framework that unifies forward and inverse problems. These studies suggest that GNN-based formulations can provide efficient and scalable models for solid mechanics.

Meanwhile, numerical methods, particularly the FEM, have been incorporated into PINN frameworks to enhance accuracy and efficiency. Several MLP-based approaches [26–28] replaced AD with FEM shape-function gradients and employed energy functionals to lower the derivative order in the loss, leading to more efficient training. In [24], coupling a graph convolutional network (GCN) with these FEM-based derivative operators was shown to better mitigate strain-localization instabilities. Another work [29] introduced an energy-based objective computed from the FEM stiffness matrix and nodal displacements. Beyond energy loss function, [30] proposed a FEM-inspired objective that evaluates nodal force residuals; compared with strong-form PINNs, it requires fewer residual terms. Wang [31] further developed a framework in which an arbitrary network backbone predicts nodal displacements, while FEM discretization is used to construct an energy loss, enabling efficient simulations with millions of degrees of freedom for elasticity and heat transfer.

In these approaches, incorporating FEM techniques and adopting GNN architectures can substantially improve the PINNs performance. However, in most cases the neural network still acts as a black-box surrogate, while FEM computations are primarily used for loss construction; the two components remain structurally decoupled. To bridge this gap, recent studies have begun to embed FEM computations directly into network design. For example, [32] represents FEM meshes as node–element hypergraphs and mimic stiffness-assembly procedures to build hypergraph neural networks for data-driven fluid dynamics modeling. Other efforts have developed differentiable FEM frameworks, which have been applied to forward simulation in elasticity and hyperelasticity [33] and to enable automated inverse design [34]. In addition, HiDeNN approximates global shape functions with neurons and built FEM-like hierarchical MLP [35] and CNN [36] architectures, achieving significant acceleration via variable-separation techniques [37,38]. Building on these advances, we introduce a physics-consistent FEM-Informed Hypergraph Neural Network (FHGNN) to improve the accuracy and efficiency of PINNs for elastoplasticity. Rather than learning a black-box mapping from spatial coordinates to displacements, FHGNN treats the displacement field as a graph attribute and embeds the core finite-element computational pipeline into message-passing operators using standard aggregation–update primitives. Compared with AD-based differentiation, we further demonstrate the sparsity and computational efficiency of derivatives evaluated via FEM shape functions. Guided by numerical experiments and Hessian conditioning analysis, we assess different loss formulations and adopt an efficient variational objective. Dirichlet boundary conditions are imposed directly through a masking vector, while Neumann conditions are naturally incorporated into the variational loss. By retaining differentiability with respect to nodal coordinates, our framework enables FEM-style r-adaptivity and attains a lower global energy than a fixed-mesh FEM baseline. We validate the method on challenging benchmarks—including 3D cantilever beams under cyclic loading and tensile tests of perforated bi-material plates with complex geometries—and perform detailed comparisons against state-of-the-art PINN variants, demonstrating superior accuracy and scalability.

The paper is organized as follows. Section 2 introduces the GNN and PINN preliminaries, and details of the proposed FHGNN. Section 3 presents numerical experiments on multiple benchmarks. Section 4 concludes with key findings and directions for future work.

## 2 Methodology

### 2.1 Graph neural networks

A general graph $\mathcal{G} = (\mathcal{V}, \mathcal{E})$ comprises features at three levels: node-level, edge-level, and graph-level features. Nodes are connected by either directed or undirected edges. Here $\mathcal{V} = \{v_1, v_2, \ldots v_n\}$ and $\mathcal{E} = \{e_{1,1}, e_{1,2}, \ldots e_{i,j}\}$ denote the sets of nodes and edges, respectively. Following the widely adopted formulation in the mainstream GNN library PyTorch Geometric (PyG), a graph neural network layer can be described as a message-passing procedure consisting of three stages: edge update, aggregation, and node update.

$$\boldsymbol{v}_i^{(k)} = \phi_v^{(k)}\left(\boldsymbol{v}_i^{(k-1)}, \oplus_{j \in \mathcal{N}(i)} \phi_e^{(k)}\left(\boldsymbol{v}_i^{(k-1)}, \boldsymbol{v}_j^{(k-1)}, \boldsymbol{e}_{j,i}\right)\right) \tag{1}$$

Where $\boldsymbol{v}_i^{(k-1)} \in \mathcal{V}$ denoting node feartures of node $i$ in the $(k-1)th$ layer and $\boldsymbol{e}_{j,i}$ denoting edge feature from node $j$ to node $i$. $\mathcal{N}(i)$ is the neighbor set of node $i$. Here $\oplus$ denotes a permutation invariant aggregation function that aggregates the information from all the edges pointing to each node $i$. $\phi_e^{(k)}$ and $\phi_v^{(k)}$ are the edge update function and node update function, respectively. If the reader is interested in more detailed definitions of specific graph convolutional kernels, we refer to the official PyTorch Geometric (PyG) documentation [39,40].

**2.2 Physics-informed neural nnetworks**

We consider a benchmark elasticity problem to demonstrate the PINN methodology. Consider a homogeneous, isotropic elastic body under small deformations, where displacement $\boldsymbol{u} = \bar{\boldsymbol{u}}$ is prescribed on the Dirichlet boundary $\varGamma_D$, and traction force $\boldsymbol{t} = \bar{\boldsymbol{t}}$ is applied on the Neumann boundary $\varGamma_N$. The strong form of the governing PDE is expressed as:

$$\begin{cases} \nabla \cdot \boldsymbol{\sigma} + \boldsymbol{f} = \boldsymbol{0}, \boldsymbol{x} \in \varOmega \\ \boldsymbol{u} = \bar{\boldsymbol{u}}, \boldsymbol{x} \in \varGamma_D \\ \boldsymbol{\sigma} \cdot \boldsymbol{n} = \bar{\boldsymbol{t}}, \boldsymbol{x} \in \varGamma_N \end{cases} \quad (2)$$

where $\boldsymbol{\sigma}$ is the Cauchy stress, $\boldsymbol{f}$ is the body force and $\boldsymbol{n}$ denotes the outward normal vector. Suppose the following isotropic linear, elastic consititutive relation can be rewritten as:

$$\boldsymbol{\sigma} = \lambda \text{tr}(\boldsymbol{\varepsilon})\boldsymbol{I} + 2\mu\boldsymbol{\varepsilon} \quad (3)$$

$$\boldsymbol{\varepsilon} = \frac{1}{2}(\nabla\boldsymbol{u} + \nabla\boldsymbol{u}^T) \quad (4)$$

where $\lambda$ and $\mu$ present the Lame constants, and can be defined by elastic modulus $E$ and Poisson's ratio $v$:

$$\lambda = \frac{vE}{(1+v)(1-2v)}, \mu = \frac{E}{2(1+v)} \quad (5)$$

Let $\mathcal{N}^L: \mathbb{R}^{D_i} \to \mathbb{R}^{D_o}$ be a fully connected feedforward neural network, with transformational relations between the adjacent layers as:

$$\mathcal{N}^k(\boldsymbol{x}) = \Phi(\boldsymbol{W}_k \mathcal{N}^{k-1}(\boldsymbol{x}) + \boldsymbol{b}_k), 1 \leq k \leq L-1 \quad (6)$$

Where $\Phi(\cdot)$ denotes the activation function, $\theta = \{\boldsymbol{W}_k, \boldsymbol{b}_k\}^{k=1,2\cdots L}$ is the set of trainable weights and bias. $L$ is the number of total layers. $\mathcal{N}^L(\boldsymbol{x})$ and $\mathcal{N}^0(\boldsymbol{x})$ are the the outputs and inputs of the network, respectively.

PINNs construct the aforementioned fully connected MLPs to approximate the displacement field $\boldsymbol{u}$. We denote the network prediction by $\mathcal{N}^L(\boldsymbol{x}) = \boldsymbol{u}^\mathcal{N}(\boldsymbol{x})$. Using AD, first-order derivatives are evaluated to obtain the strain, and the constitutive law is then applied to compute the corresponding stress prediction $\boldsymbol{\sigma}^\mathcal{N}(\boldsymbol{x})$. The equilibrium residual is defined as:

$$\boldsymbol{r}^\mathcal{N}(\boldsymbol{x}) = \nabla \cdot \boldsymbol{\sigma}^\mathcal{N}(\boldsymbol{x}) + \boldsymbol{f} \quad (7)$$

The training process minimizes a composite loss function that combines contributions

from the boundary conditions, PDE residuals, and optionally observed data:

$$\mathcal{L}_{total} = \lambda_b \mathcal{L}_b + \lambda_f \mathcal{L}_f + \lambda_d \mathcal{L}_d \tag{8}$$

Here $\lambda_b, \lambda_f, and\ \lambda_d$ are weighting coefficients for the three terms, used to balance their contributions to the gradient-based updates and to mitigate gradient imbalance caused by any single term dominating the optimization [41]. Each loss term is detailed as follows:

$$\mathcal{L}_b = \frac{1}{N_{\Gamma_D}} \sum_{i=1}^{N_{\Gamma_D}} \left\| \boldsymbol{u}^{\mathcal{N}}(\boldsymbol{x}_i^{\Gamma_D}) - \overline{\boldsymbol{u}}_i \right\|^2 + \frac{1}{N_{\Gamma_N}} \sum_{i=1}^{N_{\Gamma_N}} \left\| \boldsymbol{n}\boldsymbol{\sigma}^{\mathcal{N}}(\boldsymbol{x}_i^{\Gamma_N}) - \overline{\boldsymbol{t}}_i \right\|^2 \tag{9}$$

$$\mathcal{L}_f = \frac{1}{N_f} \sum_{i=1}^{N_f} \left\| \boldsymbol{r}^{\mathcal{N}}(\boldsymbol{x}_i^f) \right\|^2 \tag{10}$$

$$\mathcal{L}_d = \frac{1}{N_f} \sum_{i=1}^{N_d} \left\| \boldsymbol{u}^{\mathcal{N}}(\boldsymbol{x}_i^d) - \widehat{\boldsymbol{u}}(\boldsymbol{x}_i^d) \right\|^2 \tag{11}$$

Here, $\{\boldsymbol{x}_i^{\Gamma_D}\}_{i=1}^{N_{\Gamma_D}}$, $\{\boldsymbol{x}_i^{\Gamma_N}\}_{i=1}^{N_{\Gamma_N}}$, $\{\boldsymbol{x}_i^f\}_{i=1}^{N_f}$ and $\{\boldsymbol{x}_i^d\}_{i=1}^{N_d}$ denote the training points on the Dirichlet boundary, the Neumann boundary, the residual (collocation) points in the domain, and the data observation points, respectively. The quantity $\widehat{\boldsymbol{u}}(\boldsymbol{x}_i^d)$ represents the reference data at the observation points. The network parameters are trained via a gradient-based optimizer (e.g., Adam or L-BFGS) by minimizing the loss function, so that the network-defined function progressively approaches the reference displacement fields.

**2.3 FEM-Informed Hypergraph Neural Networks**

*2.3.1 Node-element Hypergraph*

In FEM meshes, connectivity is defined not only between nodes but also by the incidence between elements and their constituent nodes. In [32], a node–element hypergraph was first introduced to capture element–node incidence in FEM meshes. We adopt the same representation, $\mathcal{G} = (\mathcal{V}, \mathcal{C}, \mathcal{E})$, as shown in **Fig. 1**, where $\mathcal{V}$ denotes the set of mesh nodes, $\mathcal{C}$ represents the set of elements (treated as a second graph node type), and $\mathcal{E}$ is the incidence edges connecting each node to its associated elements (implemented as directed edges from $\mathcal{V}$ to $\mathcal{C}$). As inputs to the GNN, we construct two hypergraphs with identical topology. In $\mathcal{G}_1$, node features on $\mathcal{V}$ are the physical coordinates of mesh nodes; in $\mathcal{G}_2$, node features on $\mathcal{V}$ are the corresponding nodal displacements. $\mathcal{C}$ carry element-wise variables (typically evaluated at Gauss integration points), and edges in $\mathcal{E}$ may additionally store local incidence information and intermediate quantities required for subsequent physics-consistent message passing and updates.

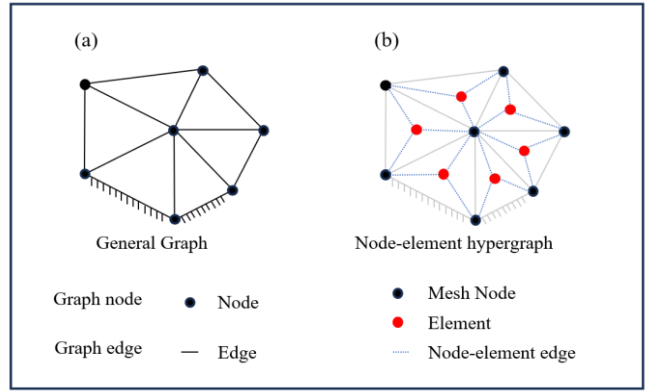

**Fig. 1.** A generic graph and the corresponding node–element hypergraph representation of an FEM mesh.

### *2.3.2 Isoparametric transformation message passing layer*

In the following derivations, we consider a linear triangular element with a single Gauss point. For brevity, the Gauss-point subscript is omitted. We first introduce an isoparametric-transformation message-passing layer $\mathcal{GNN}_1$, which updates element features and node–element edge features. Let the element nodes be indexed by $j, k, l$. Denote the natural coordinates by $\xi = (\xi, \eta)$ and the physical coordinates by $x = (x, y)$. The Jacobian matrix is given by:

$$\mathcal{J}_x(\xi) = \frac{\partial x}{\partial \xi} = \sum_j^3 \left( x_j (\nabla_\xi N_j)^T \right) \quad (12)$$

Here $N_j$ denotes the shape function associated with the $j$-th node ($j \in \{j, k, l\}$) of the element. The physical coordinates of node $j$ are $x_j = [x_j \ \ y_j]^T$, and the shape-function gradient in natural (reference) domain is $(\nabla_\xi N_j)^T = \left[ \frac{\partial N_j}{\partial \xi} \ \ \frac{\partial N_j}{\partial \eta} \right]$. For a given element type and Gauss integration points, these quantities can be precomputed as initial input features. The shape-function gradient in the physical domain is then updated as:

$$\nabla_x N_j = (\mathcal{J})^{-T} \nabla_\xi N_j \quad (13)$$

As shown in **Fig. 2**, the above computation can be readily formulated as a message-passing process:

$$c_i = \oplus_{j \in \mathcal{N}(i)} \phi_e(v_j, e_{j,i}) \quad (14)$$

$$c_i', c_i'' = \phi_v\left(\oplus_{j \in \mathcal{N}(i)} e_{j,i}'\right) \quad (15)$$

$$e''_{j,i} = \phi_e(c''_i, e_{j,i}) \qquad (16)$$

The aggregation operator is chosen to be additive. The function $\phi_e$ is realized via matrix multiplication, whereas $\phi_v$ represents the node update that evaluates the inverse and the determinant. Accordingly, we establish the following one-to-one mapping:

$$\begin{aligned} v_j &= x_j, \quad e_{j,i} = \nabla_\xi N_j \\ c_i, c'_i, c''_i &= \mathcal{J}, |\mathcal{J}|, \mathcal{J}^{-1} \\ e''_{j,i} &= \nabla_x N_j \end{aligned} \qquad (17)$$

The physical meaning of this message-passing step is as follows: given the shape-function gradients in the natural domain and the nodal coordinates of the current configuration, we compute the element Jacobian and its determinant, and use the inverse Jacobian to map the shape-function gradients from the natural domain to the physical domain. By explicitly integrating the FEM computation, this message-passing operation enables the graph neural network to learn the geometric information encoded in the isoparametric transformation.

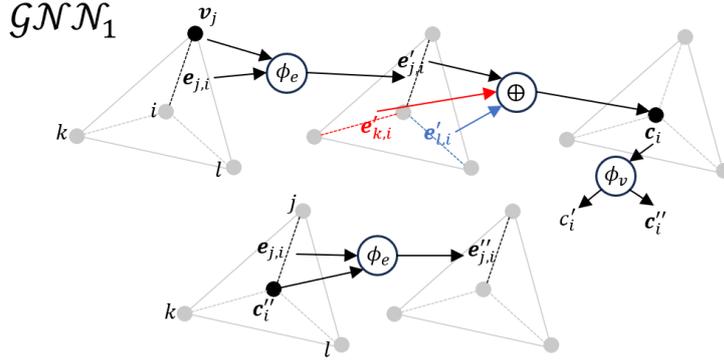

**Fig. 2.** Isoparametric transformation message-passing layer: element features are updated to include the Jacobian matrix, its determinant, and its inverse; node–element edge features are updated to obtain the physical shape-function gradients.

### 2.3.3 Strain–stress message-passing layer

We next introduce an element-wise strain–stress message-passing layer, $\mathcal{GNN}_2$. In the standard FEM setting, the strain at Gauss integration points inside each element is given by:

$$\boldsymbol{\varepsilon} = \frac{1}{2}(\nabla_x \boldsymbol{u} + (\nabla_x \boldsymbol{u})^T), \nabla_x \boldsymbol{u} = \sum_j^3 (\boldsymbol{u}_j (\nabla_x N_j)^T) \qquad (18)$$

The stress integration procedure for the elastoplastic material adopted in the following experiments is detailed in the **Appendix-**. For the purely elastic case, the stress is

updated as:

$$\boldsymbol{\sigma} = \boldsymbol{C}:\boldsymbol{\varepsilon} \tag{19}$$

Where $\boldsymbol{C}$ is the fourth-order elasticity tensor. For FHGNN, we initialize the nodal displacement $\boldsymbol{u}_j = 0$ and treat it as a trainable unkonwns. This differs from black-box graph convolutional kernels, where the trainable parameters are the weights of an underlying MLP. The node–element edge attributes $\mathcal{E}$ are initialized as $\boldsymbol{e}''_{j,i}$ (the output of $\mathcal{GNN}_1$). As illustrated in **Fig. 3**, the above procedure is implemented as another message-passing layer. Specifically, $\phi_e$ is realized via matrix multiplication and summation is adopted as the aggregation operator $\oplus$, producing the displacement gradients at Gauss points. These features are then fed into the node update function $\phi_v$ to obtain the corresponding strain $\boldsymbol{\varepsilon}_i$ and stress $\boldsymbol{\sigma}_i$:

$$\boldsymbol{\varepsilon}_i, \boldsymbol{\sigma}_i = \phi_v\left(\oplus_{j\in\mathcal{N}(i)}\phi_e\left(\boldsymbol{u}_j, \boldsymbol{e}''_{j,i}\right)\right) \tag{20}$$

By performing Gauss quadrature over the entire computational domain, we obtain the energy functional, which is one of the most popular and efficient loss functions in PINN-based solid mechanics. This functional can also be interpreted as an optimizable graph-level attribute:

$$\mathcal{L}_{energy} = \frac{1}{2}\int_\Omega \boldsymbol{\sigma}:\boldsymbol{\varepsilon}\,dV - \int_{\partial\Omega} \bar{\boldsymbol{t}}\cdot\boldsymbol{u}\,dA \tag{21}$$

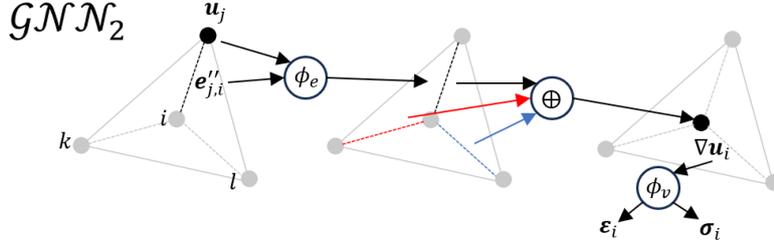

**Fig. 3.** Strain–stress message-passing layer: element features are updated to include the Gauss-point strain and stress.

### 2.3.4 Global Internal force message-passing layer

Finally, we define the internal force message-passing layer $\mathcal{GNN}_3$. In FEM, element internal forces are computed and then assembled into the global internal force vector by summing contributions at shared nodes. The element internal force vector can be computed as:

$$\boldsymbol{f}^e_j = \int_{\Omega_e} \boldsymbol{\sigma}\nabla_x N_j\,d\Omega \approx \sum_{g=1}^{n_g}\left(\boldsymbol{\sigma}_g\nabla_x N_j|_g\right)|\boldsymbol{J}_g|w_g \tag{22}$$

Where $n_g$ denotes the number of Gauss integration points, $|\mathcal{J}_g|$ is the determinant of the Jacobian, and $w_g$ is the corresponding quadrature weight. The global internal nodal force is obtained by:

$$\boldsymbol{f}_j^{int} = \sum_e \boldsymbol{f}_j^e \tag{23}$$

As illustrated in **Fig. 4**, we directly translate this procedure into the GNN framework as element-to-node message passing:

$$\boldsymbol{e}_{j,i}''' = \phi_e(\boldsymbol{\sigma}_i, \boldsymbol{e}_{j,i}'', c_i') \tag{24}$$

$$\boldsymbol{f}_j = \oplus_{i \in \mathcal{N}(j)}(\boldsymbol{e}_{j,i}''') \tag{25}$$

Eqs. (24)–(25) correspond one-to-one to Eqs. (22)–(23). Within the GNN framework, given the element–node connectivity, we realize the FEM assembly procedure through update and aggregation operations. With the global external nodal force determined by the boundary conditions, we can further construct the discrete Galerkin loss [30] as:

$$\mathcal{L}_{galerkin} = \frac{1}{N}\left\|\boldsymbol{F}^{int} - \boldsymbol{F}^{ext}\right\|_2^2 \tag{26}$$

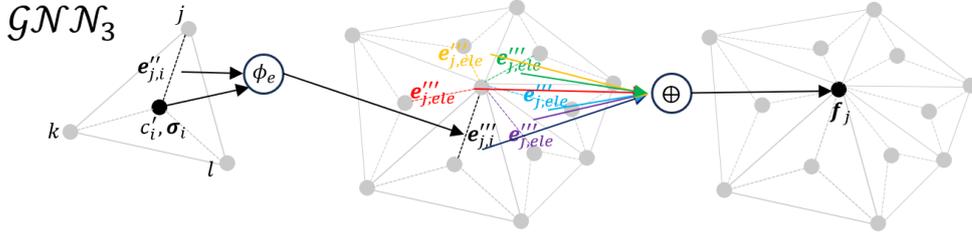

**Fig. 4.** Global internal force message-passing layer: node features are updated to include the global internal force.

### 2.3.5 Loss Function Selection and Training Procedure

The input graph attributes of FHGNN include $\boldsymbol{v}_j, \boldsymbol{e}_{j,i}$ and $\boldsymbol{u}_j$. The first two terms represent the mesh coordinates and the shape-function gradients in the natural domain, respectively, while $\boldsymbol{u}_j$ denotes the nodal displacement treated as an learnable node attribute. Through the proposed custom message-passing layers, we can compute physical quantities such as Gauss-point strain and stress, as well as the nodal internal force:

$$c_i, c_i', c_i'', \boldsymbol{e}_{j,i}'' = \mathcal{GNN}_1(\boldsymbol{v}_j, \boldsymbol{e}_{j,i}) \tag{27}$$

$$\boldsymbol{\varepsilon}_i, \boldsymbol{\sigma}_i = \mathcal{GNN}_2(\boldsymbol{u}_j, \boldsymbol{e}''_{j,i}) \tag{28}$$

$$\boldsymbol{f}_j = \mathcal{GNN}_3(\boldsymbol{c}'_i, \boldsymbol{\sigma}_i, \boldsymbol{e}''_{j,i}) \tag{29}$$

To this end, leveraging the generic node/edge attributes and the standard update–aggregation operations in the GNN framework, we embed the FEM computational pipeline directly into the message passing layers, eliminating the need to learn uninterpretable graph-based kernel parameters commonly used in related work and enabling an efficient construction of the discrete loss function. In the numerical experiments, we consider the classical $J_2$ plasticity model, whose discrete variational energy is defined in [26]. For plastic materials with a well-defined energy functional, we recommend using Eq. (21) as the loss function; detailed explanations are provided in the Section 3.6.3, where we also demonstrate that the Galerkin loss may fail to converge as the mesh is refined. When an energy functional is not readily available, Eq. (26) is adopted instead. Moreover, thanks to the end-to-end differentiable implementation, any input attribute can be treated as an optimizable parameter during training. In the subsequent experiments, we set $\boldsymbol{v}_j$ as a differentiable variable to demonstrate the mesh-adaptive capability. The procedure of FHGNN can be summarized as **Algorithm 1**.

---

**Algorithm 1** FEM-Informed Hypergraph Neural Networks.

---

**Inuput:** $\mathcal{G}_1$, $\mathcal{G}_2$ with node features $\boldsymbol{v}_j, \boldsymbol{u}_j$ and node-element edge features $\boldsymbol{e}_{j,i}$.

$\boldsymbol{u}_j.requires\_grad\_(True)$

$i = 0$

**While** $i <$ epoch **do**

   $\boldsymbol{c}_i, \boldsymbol{c}'_i, \boldsymbol{c}''_i, \boldsymbol{e}''_{j,i} = \mathcal{GNN}_1(\boldsymbol{v}_j, \boldsymbol{e}_{j,i})$

   $\boldsymbol{\varepsilon}_i, \boldsymbol{\sigma}_i = \mathcal{GNN}_2(\boldsymbol{u}_j, \boldsymbol{e}''_{j,i})$

   $\boldsymbol{f}_j = \mathcal{GNN}_3(\boldsymbol{c}'_i, \boldsymbol{\sigma}_i, \boldsymbol{e}''_{j,i})$

   If energy functional defined:

     Compute $\mathcal{L}_{energy}$

   Else:

     Compute $\mathcal{L}_{galerkin}$

   $\mathcal{L}.backward()$ and update $\boldsymbol{u}_j$

   $i \leftarrow i+1$

If update mesh: $\boldsymbol{v}_j.requires\_grad\_(True)$ and continue training

**Output:** $\boldsymbol{u}_j, \boldsymbol{\varepsilon}_i, \boldsymbol{\sigma}_i$

# 3 Results and discussion

This section presents five numerical examples using FEM solutions as references. The first 2D example compares our method with a standard PINN to highlight the difficulties of conventional PINNs in elastoplasticity. The remaining four cases benchmark FHGNN against state-of-the-art PINN variants, showing superior accuracy and efficiency. The discussion further analyzes efficiency versus FEM across mesh densities and verifies the acceleration achieved by transfer learning.

For each benchmark, FHGNN is trained and evaluated on the same mesh as the corresponding FEM model. Built upon PyG [39,40], we implement custom message-passing layers and physics-informed loss functions, enabling end-to-end differentiability of the entire framework. All neural-network computations are carried out on an NVIDIA RTX 4090 GPU server. The reference FEM solutions are generated using the commercial software Abaqus on a CPU (Intel i7-12700H, 2.70 GHz). After hyperparameter tuning, we employ the L-BFGS optimizer with an initial step size of 1.0 for all cases. For the conventional PINN baseline, the loss weights follow the recommended setting [8]: $\lambda_b = 20$ and $\lambda_f = 1$. To facilitate quantitative evaluation, we report the mean absolute error (MAE) and the relative $L_2$ error, defined as:

$$\text{MAE} = \frac{\sum_{i=1}^{N}|u_i - u_i^*|}{N} \tag{30}$$

$$L_2 = \frac{\sqrt{\sum_{i=1}^{N}|u_i - u_i^*|^2}}{\sqrt{\sum_{i=1}^{N}|u_i^*|^2}} \tag{31}$$

where $u_i$ is the predictions and $u_i^*$ denotes reference results from Abaqus.

## 3.1 2D plastic footing with cyclic loading

Here, we investigate a plastic plane-strain footing problem with the geometry shown in **Fig. 5**(a), aiming to highlight the challenges encountered by conventional PINNs in elastoplastic settings. The bottom boundary is fully fixed, while the left and right boundaries constrain the displacement in the $x$-direction. The material is characterized by an elastic modulus $E = 50$ MPa, a Poisson's ratio $\nu = 0.3$, and a perfectly plastic von Mises constitutive model with yield stress $\sigma_Y = \sqrt{3}$ MPa. A uniformly distributed cyclic load is applied on the top surface over $x = 5$–15 m.

For a conventional PINN, several independent neural networks are typically employed to predict the displacement and stress components separately. This strategy avoids second-order differentiation of the network outputs, but introduces increased memory usage, more unknown parameters, and additional loss terms. As shown in **Fig. 5(c)**, 40,000 residual training points are collected within the domain using Latin hypercube sampling [42], along with 1,200 boundary training points. The network structure comprises 14,940 trainable parameters. Data normalization is further applied to enhance numerical stability and overall performance.

For FHGNN, the loss function for the $i+1$-th load step, considering the perfect von Mises plastic model, is defined as follow:

$$Loss_{i+1} = \int_\Omega \frac{1}{2}\boldsymbol{\sigma}_{i+1}:\boldsymbol{\varepsilon}_{i+1}dV + \int_\Omega (\boldsymbol{\varepsilon}^p_{i+1} - \boldsymbol{\varepsilon}^p_i):\boldsymbol{\sigma}_{i+1}dV - \int_\Omega \bar{\boldsymbol{t}} \cdot \boldsymbol{u}_{i+1}dA \quad (32)$$

The three terms above represent the elastic strain energy, incremental dissipation, and potential energy of external loading. FHGNN employs a structured quadrilateral mesh with one Gauss point per element. The prescribed load history and mesh configuration are shown in **Fig. 6**, and the analysis is carried out in four load steps. The results from each load step, including network parameters, graph attributes, and plastic internal variables, are carried forward to initialize the subsequent step. **Fig. 7** presents the predictions of $u_x$ at each load step, together with the pointwise absolute errors of FHGNN and the conventional PINN, where the FEM solution is used as the reference. In the first load step, the external load induces a small amount of plastic deformation, resulting in displacement fields close to the elastic response; consequently, the PINN achieves acceptable accuracy. In the second load step (unloading), no additional plastic deformation develops, and the conventional PINN also yields reasonable predictions. However, in the third load step, additional plastic deformation occurs during loading, and the accuracy of the conventional PINN deteriorates markedly. By the fourth load step, the accumulated errors cause the conventional PINN predictions to diverge significantly from the reference solution. In contrast, the proposed FHGNN consistently delivers accurate predictions across all four load steps. **Fig. 8** shows the results for the equivalent plastic strain, which is defined as:

$$\bar{\varepsilon}^p = \sqrt{\frac{2}{3}\boldsymbol{\varepsilon}^p:\boldsymbol{\varepsilon}^p} \quad (33)$$

where evaluations and comparisons are conducted at the Gauss points. **Fig. 8** demonstrates that, although the conventional PINN provides similar results for displacements in step 1, the predicted equivalent plastic strains exhibit significant errors. Capturing the nonlinear response of plastic materials remains a challenge for the conventional PINN. Furthermore, to accurately assess plasticity predictions, the focus should be on the plastic internal variables rather than solely on the displacements. A more detailed comparison of training results is presented in **Table 1**. Compared with the conventional PINN, whose accuracy progressively deteriorates, FHGNN achieves an over 10 times speedup while maintaining accurate predictions.

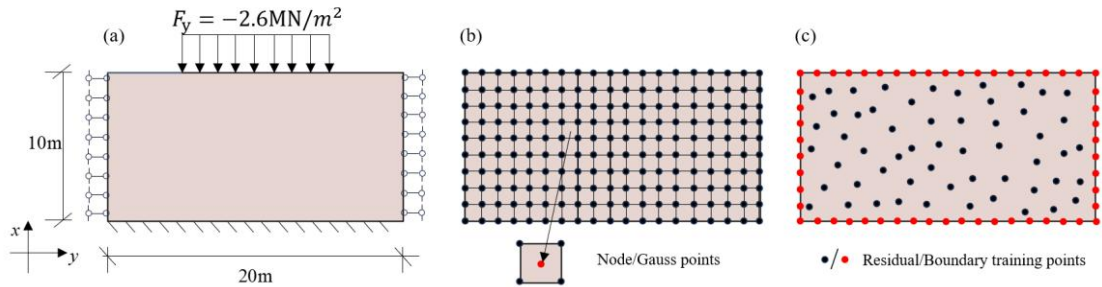

**Fig. 5.** 2D plastic footing: (a) Geometry and boundary conditions; (b) FEM mesh and Gaussian point; (c) PINN training point.

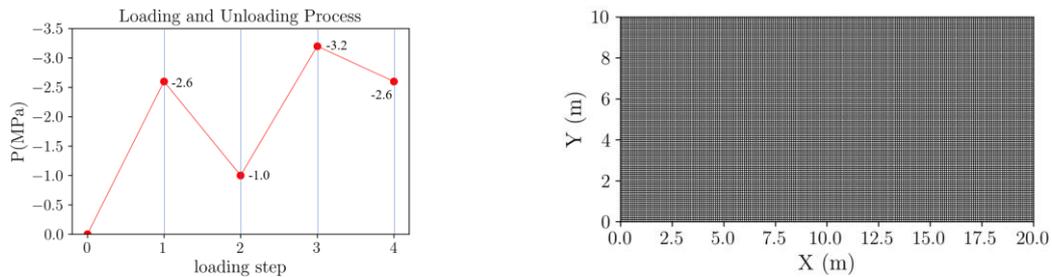

**Fig. 6.** 2D plastic footing: Cyclic loading history and schematic of the mesh used for FHGNN.

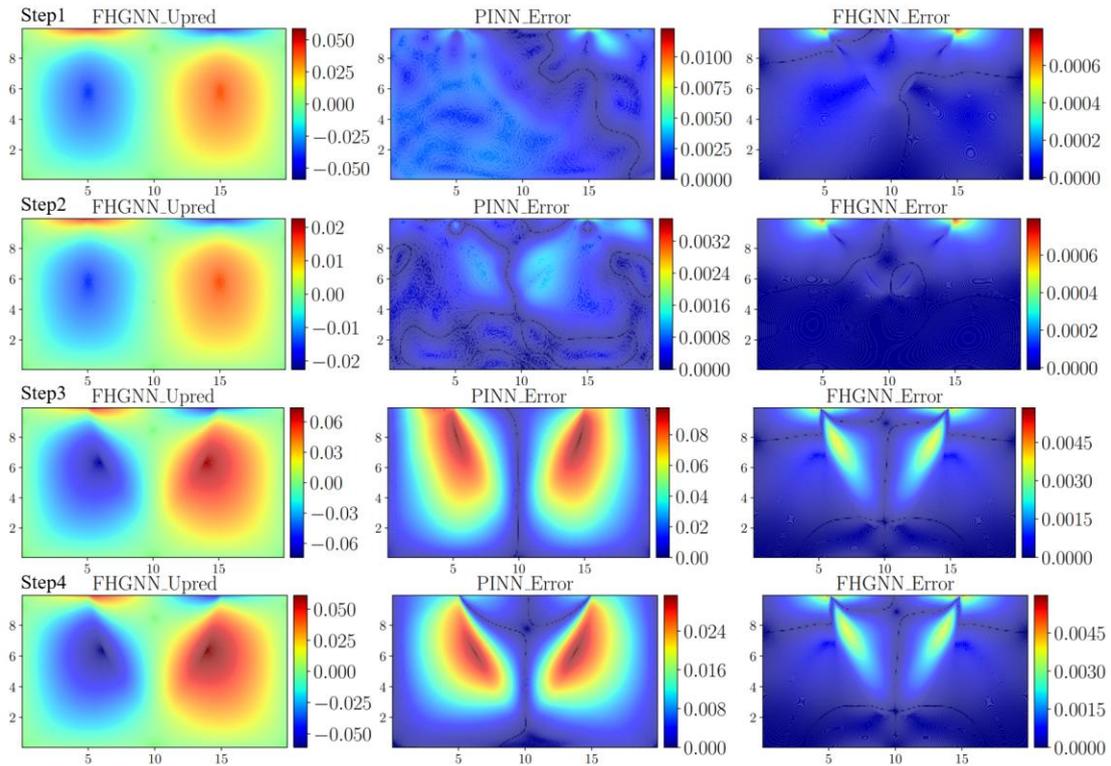

**Fig. 7.** 2D plastic footing: FHGNN's prediction, conventional PINN's and FHGNN's pointwise absolute error for $u_x$ (load steps 1–4 shown from top to bottom).

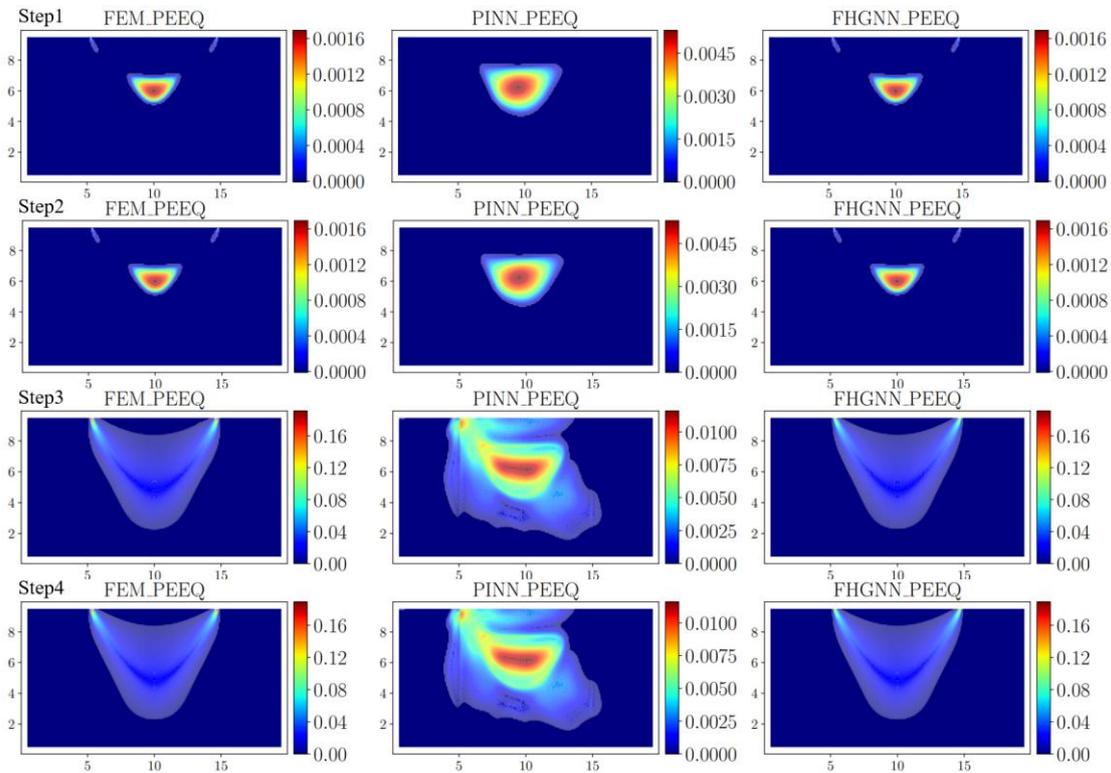

**Fig. 8.** 2D plastic footing: Evolution of the equivalent plastic strain predicted by FEM, conventional PINN and FHGNN (load steps 1–4 shown from top to bottom).

**Table 1**

Performance comparison of conventional PINN and FHGNN at four load steps.

| 10 | Method | Time(s) | $L_2\_u_x$ | $L_2\_u_y$ | $MAE\_\bar{\varepsilon}^p$ |
|---|---|---|---|---|---|
| 1 | PINN | 402 | 5.91E-2 | 1.04E-2 | 8.81E-4 |
|   | FHGNN | 36 | 3.79E-3 | 1.03E-3 | 3.89E-6 |
| 2 | PINN | 342 | 5.74E-1 | 5.92E-1 | 1.81E-4 |
|   | FHGNN | 23 | 9.58E-3 | 1.30E-3 | 3.89E-6 |
| 3 | PINN | 352 | 1.76E+0 | 6.12E-1 | 3.45E-3 |
|   | FHGNN | 31 | 2.74E-2 | 8.10E-3 | 4.82E-4 |
| 4 | PINN | 351 | 1.48E-1 | 1.74E-1 | 3.45E-3 |
|   | FHGNN | 43 | 3.10E-2 | 9.80E-3 | 4.82E-4 |

### 3.2 3D Plastic footing

Here we consider a 3D plastic footing problem. The geometry follows **Fig. 5(a)**, with a thickness of 1 m along the $z-$axis. A uniformly distributed pressure $P = -3.2$ MPa is applied on the top surface over $y \in [5,15]\ m$, using the same plastic material as described in Section 3.1. The bottom is fully fixed, displacements along the $z$-axis are constrained to zero, and displacements in the x-direction are fixed on the left and right boundaries. For the conventional PINNs, the additional dimension leads to a substantial increase in the number of residual training points. Prior studies have shown that PINNs struggle to accurately capture plastic responses even in 2D, and our experiments further indicate that their performance in 3D cases is even poorer. Therefore, a comparison with the conventional PINN is unnecessary; instead, we benchmark FHGNN against other state-of-the-art physics-driven training approaches. In several recent studies [26,31], an MLP is used to predict nodal displacements and a variational loss is constructed using FEM shape-function gradients; we refer to this class of methods as PIMLP. In contrast, other works [24,25] investigate the representation capability of classical GNNs by learning a nonlinear mapping from nodal coordinates to nodal displacements on a given mesh; we collectively denote these methods as PIGCN. For a fair comparison, all subsequent experiments follow the settings in [24]. Specifically, PIGCN adopts a Chebyshev spectral graph convolution operator [20] with one-hop neighborhoods per layer, while PIMLP and PIGCN use comparable network architectures with layer widths [3,16,32,64,32,16,3].

Both Abaqus and our FHGNN utilize linear hexahedral elements here. A relatively dense mesh is employed, resulting in 182,709 degrees of freedom (DOFs). We use the

L-BFGS optimizer with an initial step size of 1.0 and a "strong Wolfe" line search. As shown in **Fig. 9** and **Table 2**, FHGNN converges to high-accuracy predictions after about 1000 training iterations, taking 125 s, with final relative $L_2$ errors of 8.06E-03 and 2.40E-03 for the $u_x$ and $u_y$ displacement components, respectively. Compared with PIGCN and PIMLP, the custom message passing in FHGNN yields faster convergence and more accurate predictions. **Fig. 10** compares the predictions from FHGNN with the FEM results, demonstrating that FHGNN can accurately solve 3D plasticity problems.

For FHGNN, the graph attributes used as inputs include $v_j$, $e_{j,i}$, and $u_j$. The FHGNN framework is end-to-end differentiable: all input quantities can be implemented as PyTorch leaf tensors, enabling backpropagation through the entire workflow. By treating the nodal coordinates $v_j$ as differentiable variables, FHGNN exhibits behavior analogous to $r$-adaptive mesh updates in FEM. Using the 3D plastic footing example in this section, we keep all other settings unchanged and adopt a coarse mesh for clearer visualization. The problem is first solved on a uniform mesh. After 4,000 training iterations, the predicted $x$-direction displacement $u_x$ is shown in **Fig. 11(a)**, with the energy loss converging to $-6.322$, which matches the potential energy computed by FEM solution on the fixed mesh. We then treat the mesh coordinates as differentiable leaf tensors, and the gradients of the loss with respect to the coordinates are computed for mesh updates. To prevent mesh penetration or drifting outside the computational domain, all DOFs at boundary nodes are fixed, and the $z$-coordinates are fixed for all nodes. **Fig. 11(c)** shows the predictions after an additional 1,000 optimization steps, where the mesh is automatically refined in regions with sharp gradients and becomes coarser in smoother areas. The system potential energy is further reduced to $-6.328$. Moreover, the proposed framework also enables the optimization of Gauss points, which we will explore in future work to obtain accurate solutions with fewer integration points.

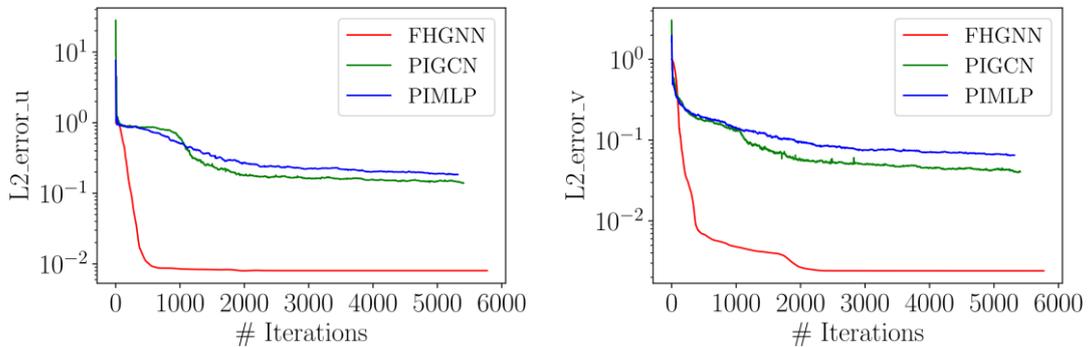

**Fig. 9.** 3D plastic footing: Training histories of the relative $L_2$ errors of $u_x$ and $u_y$ for FHGNN, PIGCN, and PIMLP.

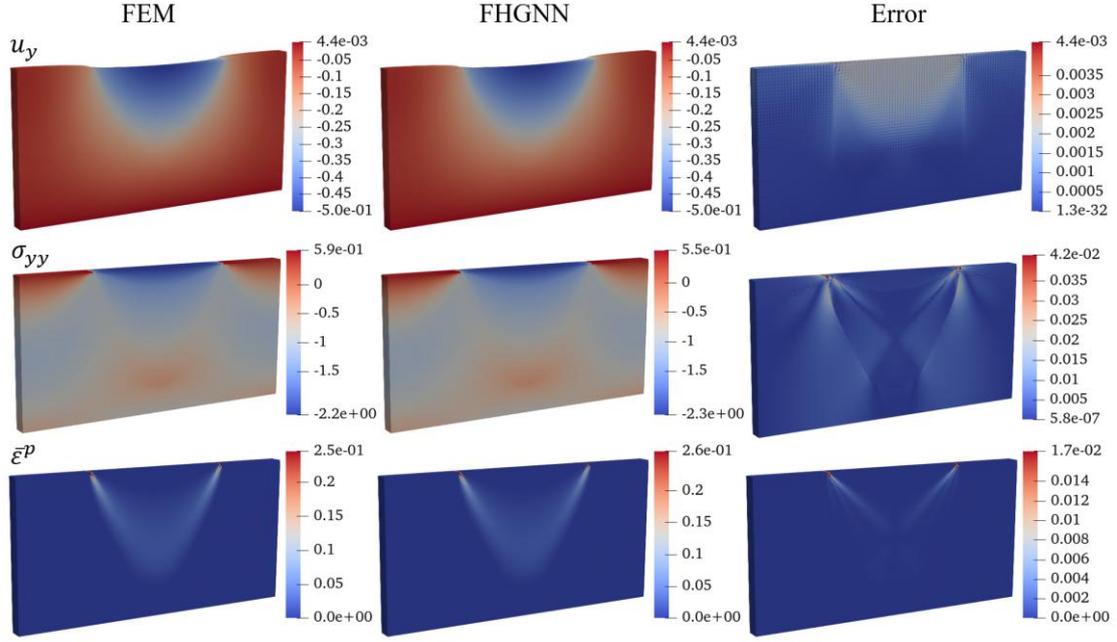

**Fig. 10.** 3D plastic footing: Abaqus reference solution (left column), FHGNN's predictions (middle) and pointwise absolute error (right column) for $u_y$, $\sigma_{yy}$ and $\bar{\varepsilon}^p$ (from top to bottom row).

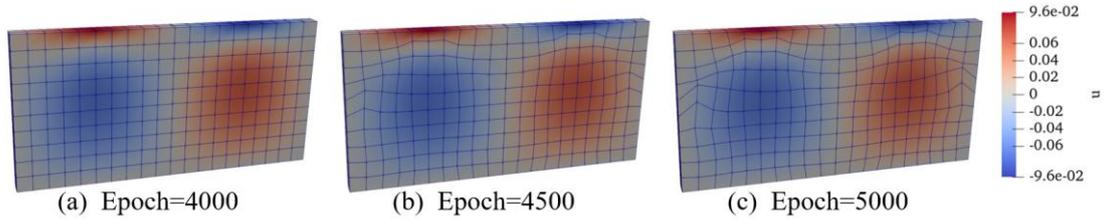

(a) Epoch=4000  (b) Epoch=4500  (c) Epoch=5000

**Fig. 11.** Different mesh distribution and predictions after updating mesh automatically through FHGNN.

**Table 2**
Performance comparison of PIMLP, PIGCN, and FHGNN on the 3D plastic footing problem.

| Method | Time(s) | $L_2\_u_x$ | $L_2\_u_y$ | $MAE\_\bar{\varepsilon}^p$ |
| --- | --- | --- | --- | --- |
| PIMLP | 210.47 | 1.85E-01 | 6.52E-02 | 2.50E-3 |
| PIGCN | 452.08 | 1.28E-01 | 3.38E-02 | 2.10E-3 |
| FHGNN | 125.23 | 8.06E-03 | 2.40E-03 | 8.62E-05 |

### 3.3 3D linear hardening cantilever beam

In this section, we analyze a 3D cantilever beam composed of a linear hardening plastic material. We evaluate both linear isotropic and kinematic hardening plasticity models, under which the yield surface respectively expands or translates during plastic flow. The material parameters considered in this section include an elastic modulus of $E = 200$ MPa, a Poisson's ratio of 0.3, an initial yield stress of $\sqrt{3}$ MPa, a linear isotropic hardening modulus of $K = 100$ MPa, and a linear kinematic hardening modulus of $H = 100$ MPa. **Fig. 12** illustrates the characteristics of these two different hardening models.

The geometry and boundary conditions of the cantilever beam are shown in **Fig. 15(a)**. The beam has a length of 4 m in the $x$-direction and a cross-sectional width of 1 m in the $y$–$z$ plane. The left end is fully fixed, while a prescribed displacement of $-0.25$ m in the $y$-direction is applied at the right end. The mesh consists of $160 \times 40 \times 40$ 8-node linear hexahedral elements, resulting in 768,000 DOFs, yielding a relatively large input graph for GNN-based models. Such a fine mesh facilitates accurate resolution of the nonlinear response within the plastic zone. We consider a $J_2$ plasticity model with a linear isotropic hardening modulus $K = 100$ MPa. Following the classical discrete variational principle of elastoplasticity [43], the energy functional is defined as the sum of the total free energy and the incremental dissipation:

$$P_{t+1} = \int_\Omega \begin{pmatrix} W_{t+1} + \frac{1}{2} \boldsymbol{v}_{t+1} \cdot \boldsymbol{D}^{-1} \boldsymbol{v}_{t+1} - \Delta\gamma f_{t+1} + \\ (\boldsymbol{\varepsilon}^p_{t+1} - \boldsymbol{\varepsilon}^p_t) : \boldsymbol{\sigma}_{t+1} - \boldsymbol{v}_{t+1} \cdot \boldsymbol{D}^{-1}(\boldsymbol{v}_{t+1} - \boldsymbol{v}_t) \end{pmatrix} dV - \int_\Omega \bar{\boldsymbol{t}} \cdot \boldsymbol{u}_{i+1} dA \quad (34)$$

where $W_{t+1}$ is the elastic strain energy. $\boldsymbol{v}$ and $\boldsymbol{D}$ denote the collection of internal plastic variables and matrix of hardening moduli:

$$\boldsymbol{v}_{t+1} = \begin{bmatrix} K\bar{\varepsilon}^p \\ \sqrt{\frac{3}{2}} \boldsymbol{q}_{t+1} \end{bmatrix} \quad (35)$$

$$\boldsymbol{D} = \begin{bmatrix} K & 0 \\ 0 & H\boldsymbol{I}_{[3\times 3]} \end{bmatrix} \quad (36)$$

Here $\boldsymbol{I}$ is a second-order identity tensor. The plastic multiplier in the $J_2$ plasticity model admit closed-form expressions. By employing the radial return mapping method, the discrete consistency conditions and the KKT conditions are satisfied by construction. Consequently, the energy loss function for the case of linear isotropic hardening is defined as:

$$Loss = \int_\Omega (W_{t+1} + \frac{1}{2} K(\bar{\varepsilon}^p_{t+1})^2 + (\boldsymbol{\varepsilon}^p_{t+1} - \boldsymbol{\varepsilon}^p_t) : \boldsymbol{\sigma}_{t+1} - K\bar{\varepsilon}^p_{t+1}(\bar{\varepsilon}^p_{t+1} - \bar{\varepsilon}^p_t)) dV - P_{ext} \quad (37)$$

The reference solution is computed using Abaqus with parallel execution on 8 CPU cores, requiring 399 s. The proposed FHGNN converges after 1,844 iterations (we use the L-BFGS optimizer with a preset budget of 5,000 iterations for all methods, while FHGNN converges earlier), taking 145.51 s and achieving an approximately 3 × speedup over FEM. The final relative $L_2$ errors are 4.800E-04, 3.194E-04, and 3.646E-03 for the $x$-, $y$-, and $z$-displacement components, respectively. In contrast, PIGCN and PIMLP require 2,520.6 s and 737.6 s, respectively. **Fig. 13** compares the training histories of the displacement relative $L_2$ errors for FHGNN, PIGCN, and PIMLP. **Fig. 14** further presents a detailed comparison of the predicted fields $u_y$, $\varepsilon_{yy}^p$, and $\bar{\varepsilon}^p$. FHGNN achieves the highest accuracy and captures the stress concentration near the fixed boundary more faithfully. PIGCN also outperforms PIMLP, further indicating that GNN-based architectures are better suited for this class of problems. As summarized in **Table 3**, the predicted plastic internal variables from FHGNN are highly accurate, with relative $L_2$ errors below 3% and MAE less than 6.2E-6.

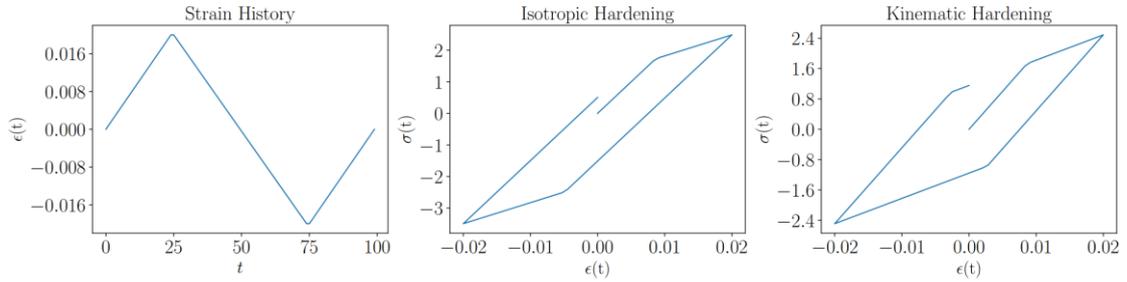

**Fig. 12.** Stress-strain curves for linear isotropic/kinematic hardening under given strain history.

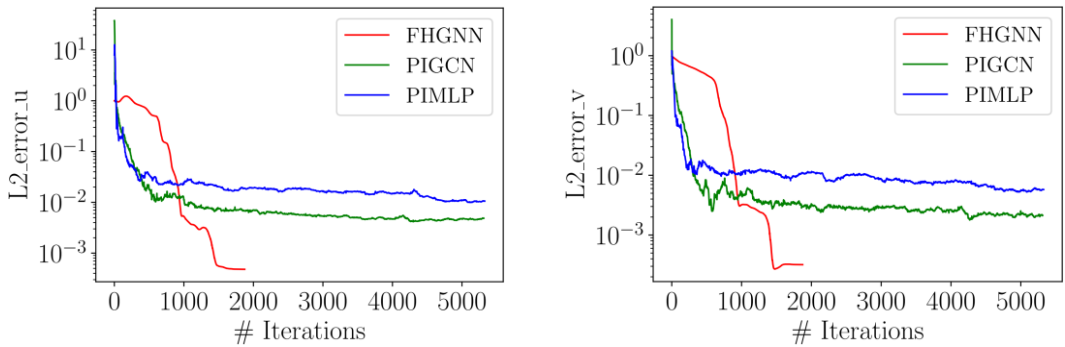

**Fig. 13.** 3D linear isotropic hardening beam: Training histories of the relative $L_2$ errors of $u_x$ and $u_y$ for FHGNN, PIGCN, and PIMLP.

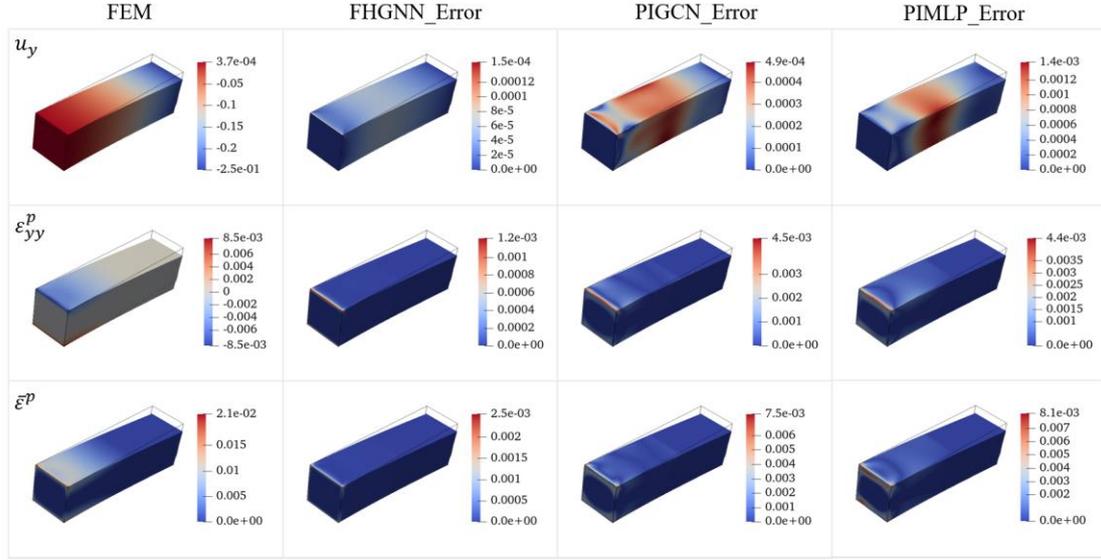

**Fig. 14.** 3D linear isotropic hardening cantilever beam: Abaqus reference solution and pointwise absolute errors of FHGNN, PIGCN, and PIMLP for $u_y$, $\varepsilon_{yy}^p$ and $\bar{\varepsilon}^p$ (from top to bottom row).

**Table 3**

Error indices of FHGNN on the 3D linear isotropic hardening cantilever beam problem.

| Variable | $L_2$ Error | MAE |
| --- | --- | --- |
| $\sigma_{xx}$ | 2.377E-03 | 9.466E-4 |
| $\sigma_{yy}$ | 6.025E-02 | 6.366E-4 |
| $\sigma_{zz}$ | 3.514E-02 | 6.725E-4 |
| $\varepsilon_{xx}^p$ | 1.893E-02 | 5.843e-6 |
| $\varepsilon_{yy}^p$ | 2.626E-02 | 3.460e-6 |
| $\varepsilon_{zz}^p$ | 1.977E-02 | 2.598e-6 |
| $\bar{\varepsilon}^p$ | 1.992E-02 | 6.103e-6 |

The cantilever beam is further extended to a length of 6 m in the $x$-direction, and three load steps are applied sequentially at the right end, with prescribed vertical displacements of -0.5 m, 0.8 m, and -1.2 m. A linear kinematic hardening material model is considered. FHGNN and FEM employ the same mesh configuration, which contains 22,527 DOFs. By replacing the hardening term in Eq. (34) with the kinematic hardening modulus, the energy loss function for the linear kinematic hardening case is derived as follows:

$$Loss = \int_\Omega (W_{t+1} + \frac{3}{4H}\boldsymbol{q}_{t+1}:\boldsymbol{q}_{t+1} + (\boldsymbol{\varepsilon}_{t+1}^p - \boldsymbol{\varepsilon}_t^p):\boldsymbol{\sigma}_{t+1} - \frac{3}{2H}\boldsymbol{q}_{t+1}:(\boldsymbol{q}_{t+1}-\boldsymbol{q}_t))dV - P_{ext} \quad (38)$$

During training, the converged nodal displacements from the previous load step are used to initialize the subsequent step. This strategy leverages transfer learning across load steps,, and the resulting loss histories of FHGNN are shown in **Fig. 15(b)**. For each load step, all three methods are trained for 5,000 iterations using the L-BFGS optimizer. **Fig. 16**, **Fig. 17** and **Table 4** provide detailed comparisons of the predicted equivalent plastic strain and equivalent von Mises stress under cyclic loading, where the von Mises stress is defined as:

$$\bar{\sigma} = \sqrt{\frac{3}{2}\boldsymbol{\sigma}':\boldsymbol{\sigma}'} \tag{39}$$

Compared with PIGCN and PIMLP, FHGNN achieves more than a 3× speedup, while reducing the average prediction error by approximately two orders of magnitude.

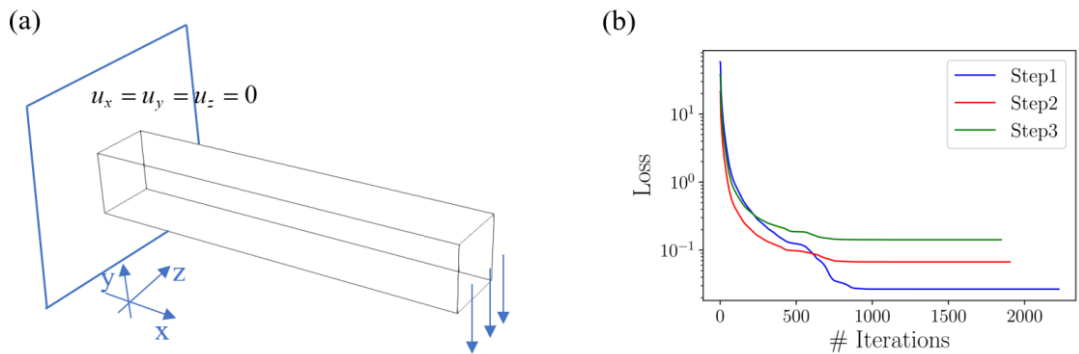

**Fig. 15.** 3D linear hardening cantilever beam: (a) Geometry and boundary conditions; (b) Training loss histories at different load steps in linear kinematic hardening case.

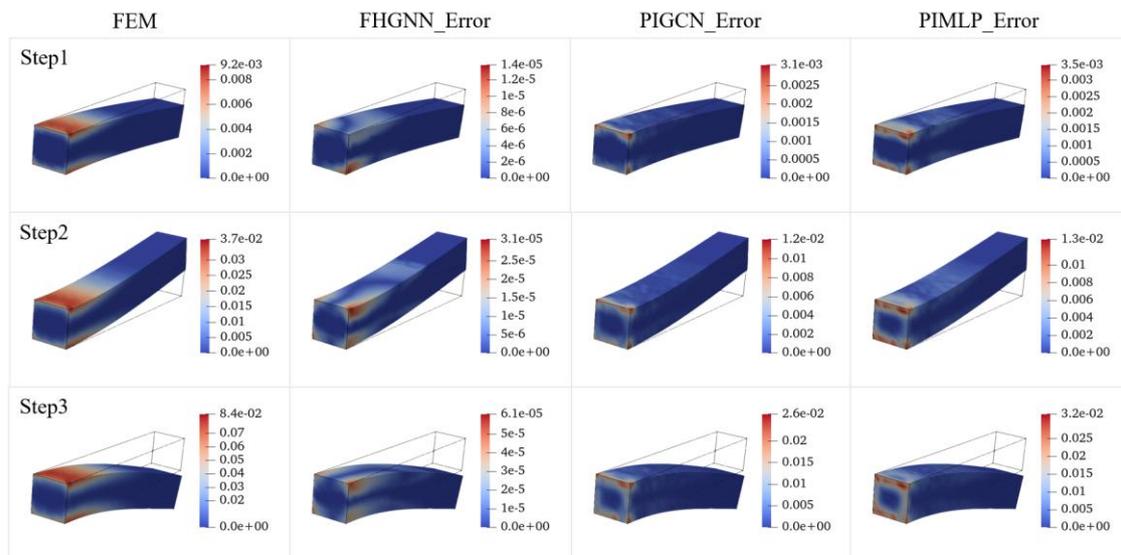

**Fig. 16.** 3D linear kinematic hardening cantilever beam under cyclic loading: Abaqus reference solution and pointwise absolute errors of FHGNN, PIGCN, and PIMLP for $\bar{\varepsilon}^p$ at three different load steps (from top to bottom row).

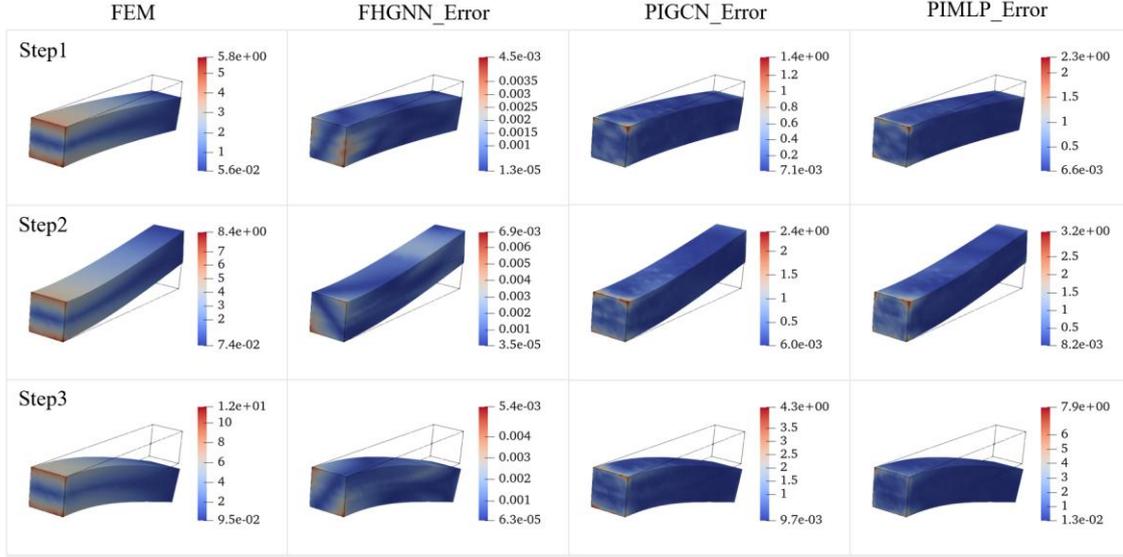

**Fig. 17.** 3D linear kinematic hardening cantilever beam under cyclic loading: Abaqus reference solution and pointwise absolute errors of FHGNN, PIGCN, and PIMLP for $\bar{\sigma}$ at three different load steps (from top to bottom row).

**Table 4**
Performance comparison of PIMLP, PIGCN, and FHGNN on the 3D linear kinematic hardening cantilever beam problem. (The best-performing results are highlighted in red.)

| Step | Method | Time(s) | $L_2\_\bar{\sigma}$ | $L_2\_\bar{\varepsilon}^p$ | MAE$\_\bar{\sigma}$ | MAE$\_\bar{\varepsilon}^p$ |
|---|---|---|---|---|---|---|
| 1 | PIMLP | 96.994 | 6.945E-02 | 1.614E-01 | 6.185E-02 | 9.109E-05 |
|   | PIGCN | 175.132 | 5.743E-02 | 1.027E-01 | 5.131E-02 | 5.923E-05 |
|   | FHGNN | 32.720 | 3.215E-04 | 8.612E-04 | 3.176E-04 | 5.914E-07 |
| 2 | PIMLP | 99.002 | 9.903E-02 | 1.374E-01 | 1.105E-01 | 3.898E-04 |
|   | PIGCN | 178.083 | 7.966E-02 | 9.100E-02 | 9.091E-02 | 2.675E-04 |
|   | FHGNN | 27.989 | 5.288E-04 | 6.066E-04 | 8.512E-04 | 2.278E-06 |
| 3 | PIMLP | 102.122 | 1.058E-01 | 1.173E-01 | 1.512E-01 | 9.330E-04 |
|   | PIGCN | 178.084 | 9.910E-02 | 8.554-E-02 | 1.410E-01 | 7.235E-04 |
|   | FHGNN | 28.805 | 3.081E-04 | 4.712E-04 | 6.483E-04 | 5.009E-06 |

### 3.4 3D linear hardening workpiece

In this case, we consider a 3D workpiece with a complex geometry. For PINNs, solving such problems is particularly challenging, as geometric complexity typically demands substantially more residual (collocation) points and exacerbates the pathological imbalance among gradients from multiple loss terms [18]. By contrast,

FHGNN leverages discrete graph attributes to directly impose Dirichlet boundary conditions, making boundary enforcement straightforward:

$$\boldsymbol{u}_j = \boldsymbol{u}^{var} \odot \boldsymbol{m} + \bar{\boldsymbol{u}} \tag{40}$$

where $\boldsymbol{u}^{var}$ is defined as a differentiable leaf tensor, $\odot$ denotes the Hadamard product and $\boldsymbol{m}$ is a binary mask with zeros at Dirichlet boundary DOFs and ones elsewhere. $\bar{\boldsymbol{u}}$ stores the prescribed displacement boundary values at the corresponding Dirichlet DOFs and zeros otherwise. With this design, the constructed displacement vector $\boldsymbol{u}_j$ fed into $\mathcal{GNN}_2$ always satisfies the Dirichlet boundary conditions, regardless of updates to $\boldsymbol{u}^{var}$. Compared with MLP-based PINNs, this discrete strategy readily handles non-trivial boundary conditions without requiring auxiliary functions. **Fig. 18** illustrates the geometry, boundary conditions, and mesh of the problem. The base of the workpiece is fully fixed, while a vertical displacement of 0.5 cm is applied to the right arm. The material parameters are defined as follows: elastic modulus $E = 2000$ MPa, Poisson's ratio $v = 0.3$, initial yield stress $\sigma_Y = \sqrt{3}$ MPa, and linear kinematic hardening modulus $H = 100$ MPa. The mesh consists of 33,367 elements, resulting in a total of 22,521 DOFs. All methods are trained using the L-BFGS optimizer for 5,000 iterations, and the final predictions are reported in **Fig. 19**, **Fig. 20** and **Table 5**. The predictions obtained from the proposed method are highly consistent with the FEM results. In particular, FHGNN successfully captures the plastic response along both sides of the arm and near its root, whereas PIGCN and PIMLP fail to provide physically reasonable predictions.

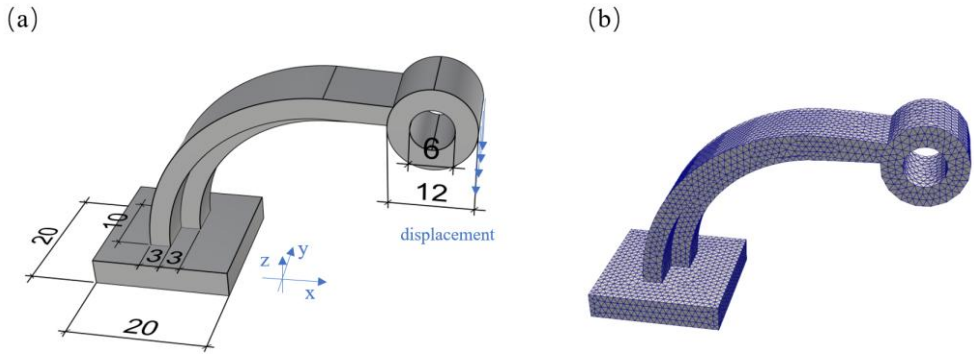

**Fig. 18.** 3D linear kinematic hardening workpiece: (a) Geometry and boundary conditions; (b) Mesh configuration.

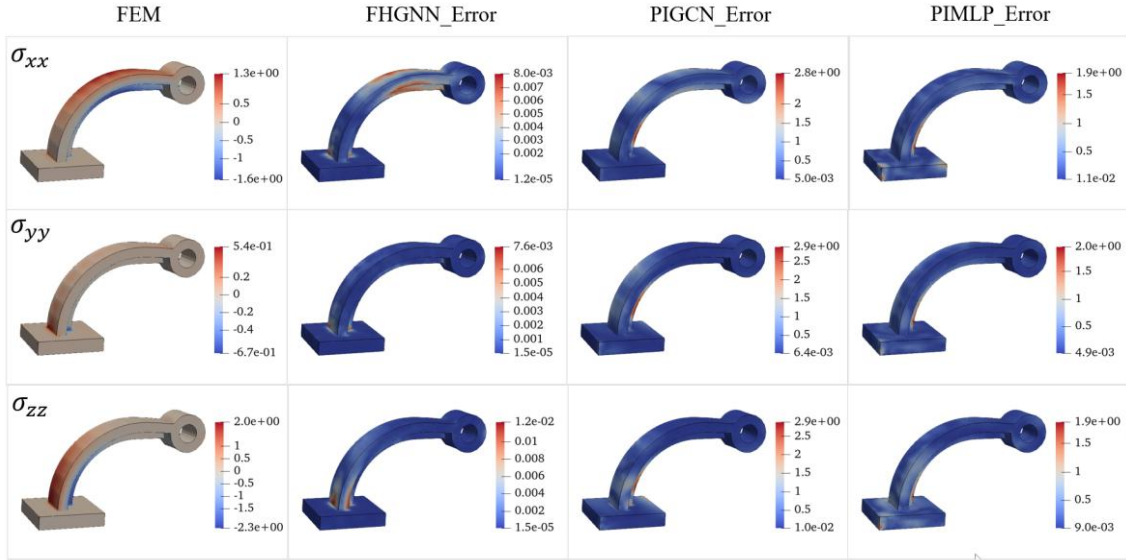

**Fig. 19.** 3D linear kinematic hardening workpiece: Abaqus reference solution and pointwise absolute errors of FHGNN, PIGCN, and PIMLP for $\sigma_{xx}$, $\sigma_{yy}$ and $\sigma_{zz}$ (from top to bottom row).

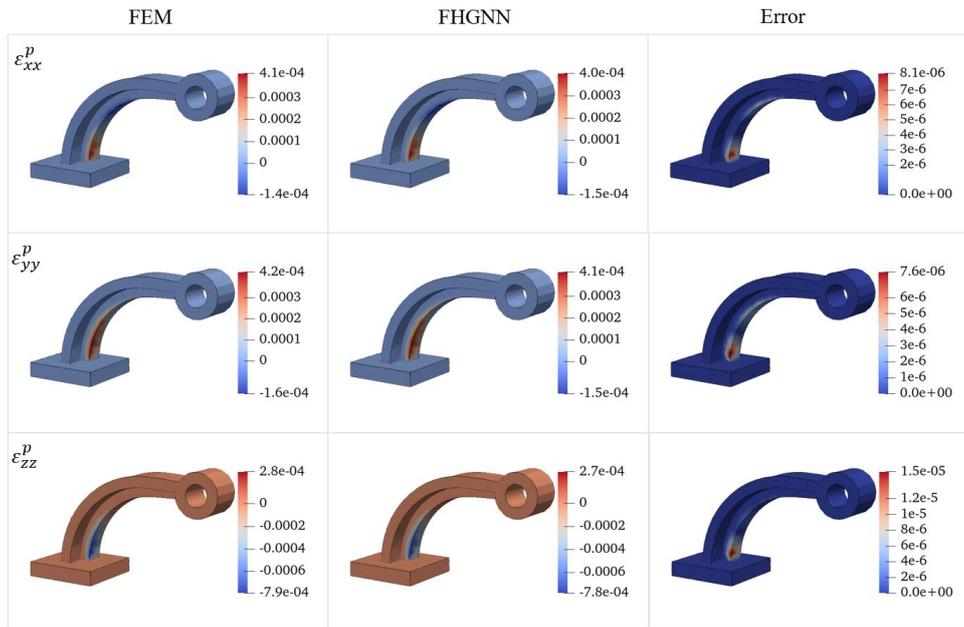

**Fig. 20.** 3D linear kinematic hardening workpiece: Abaqus reference solution (left column), FHGNN' predictions (middle) and pointwise error (right column) for $\varepsilon^p_{xx}$, $\varepsilon^p_{yy}$ and $\varepsilon^p_{zz}$ (from top to bottom row).

**Table 5**
Performance comparison of PIMLP, PIGCN, and FHGNN on the 3D linear hardening workpiece problem.

| Method | Time(s) | $L_2$ Error | | | | MAE | | | |
|---|---|---|---|---|---|---|---|---|---|
| | | $\sigma_{xx}$ | $\sigma_{yy}$ | $\sigma_{zz}$ | $\bar{\varepsilon}^p$ | $\sigma_{xx}$ | $\sigma_{yy}$ | $\sigma_{zz}$ | $\bar{\varepsilon}^p$ |
| PIMLP | 98.42 | 6.49E-01 | 2.60E+00 | 5.50E-01 | 8.02E-01 | 1.61E-01 | 1.26E-01 | 1.64E-01 | 1.02E-05 |
| PIGCN | 169.71 | 1.13E+00 | 4.54E+00 | 8.44E-01 | 1.92E+00 | 2.61E-01 | 1.79E-01 | 2.02E-01 | 2.94E-05 |
| FHGNN | 52.13 | 5.18E-03 | 8.61E-03 | 4.03E-03 | 1.77E-02 | 1.13E-03 | 3.31E-04 | 8.74E-04 | 1.90E-07 |

### 3.5 3D bi-material plate with a hole

The above examples highlight the advantages of FHGNN in handling complex geometry. Here, we further consider a perforated plate composed of two materials to demonstrate that FHGNN can efficiently capture the localized high-gradient fields induced by material discontinuities. A domain-decomposition PINN has been proposed in [7], but it requires multiple sub-networks and additional interface loss terms, thereby increasing the computational cost. In contrast, FHGNN is built on a node–element hypergraph: material information is naturally associated with each element and is directly used to update Gauss-point stresses, without introducing any extra interface loss terms. As shown in **Fig. 21**, we consider a rectangular plate of size $8 \times 8 \times 1 \text{ m}^3$, with a central circular hole of radius 1.5 m. The left boundary is fully fixed, and a horizontal displacement of 0.5 m is prescribed on the right side. The material properties are as follows: for material 1, $E_1 = 200$ MPa, $v_1 = 0.3$, $\sigma_{Y1} = \sqrt{3}$ MPa, and $H_1 = 100$ MPa; for material 2, $E_2 = 150$ MPa, $v_2 = 0.3$, $\sigma_{Y2} = \sqrt{3}$ MPa, and $H_2 = 50$ MPa. The model is discretized using linear tetrahedral elements. The unstructured mesh contains 54,402 elements and 33,711 DOFs.

FHGNN, PIGCN, and PIMLP are each trained for 5,000 iterations, and the corresponding results are summarized in **Fig. 22**, **Fig. 23** and **Table 6**. Although **Fig. 22** shows that PIGCN and PIMLP yield reasonably accurate displacement fields (with relative errors on the order of 1E-02), **Table 6** indicates that they fail to reliably predict stresses or plastic strains. This is because displacements are the direct network outputs, whereas stresses, strains, and plastic strains are derived from displacement gradients; consequently, accuracy does not consistently carry over to the first-derivative fields. In contrast, FHGNN attains the shortest runtime (59.8 s) and accurately predicts the stress–strain fields under material heterogeneity, while faithfully capturing the sharp solution features near the fixed boundary and around the circular hole.

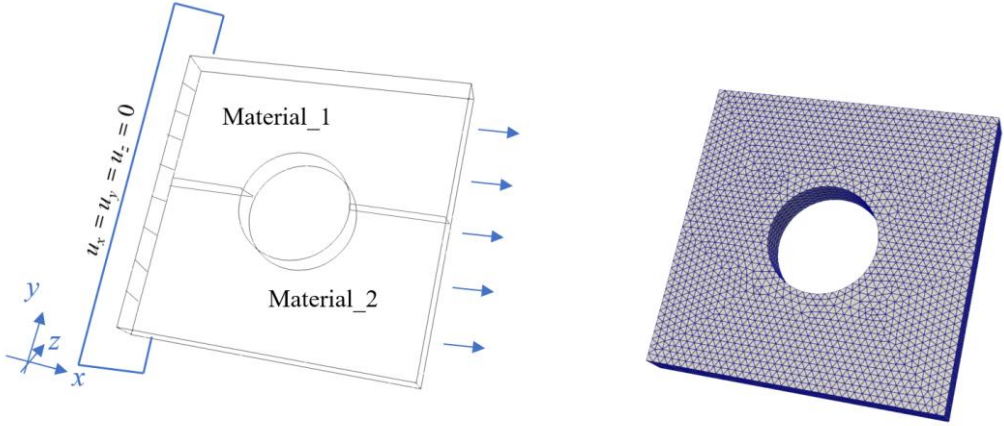

**Fig. 21.** 3D bi-material plate with a hole: (a) Geometry and boundary conditions; (b) Mesh configuration.

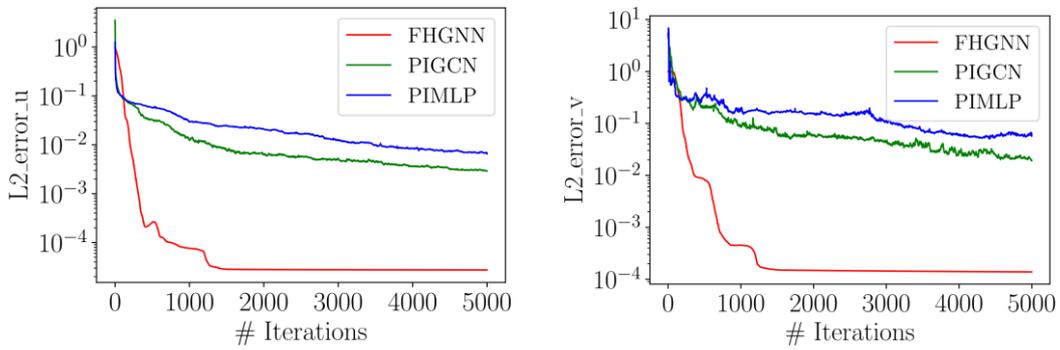

**Fig. 22.** 3D bi-material plate with a hole: Training histories of the relative $L_2$ errors of $u_x$ and $u_y$ for FHGNN, PIGCN, and PIMLP.

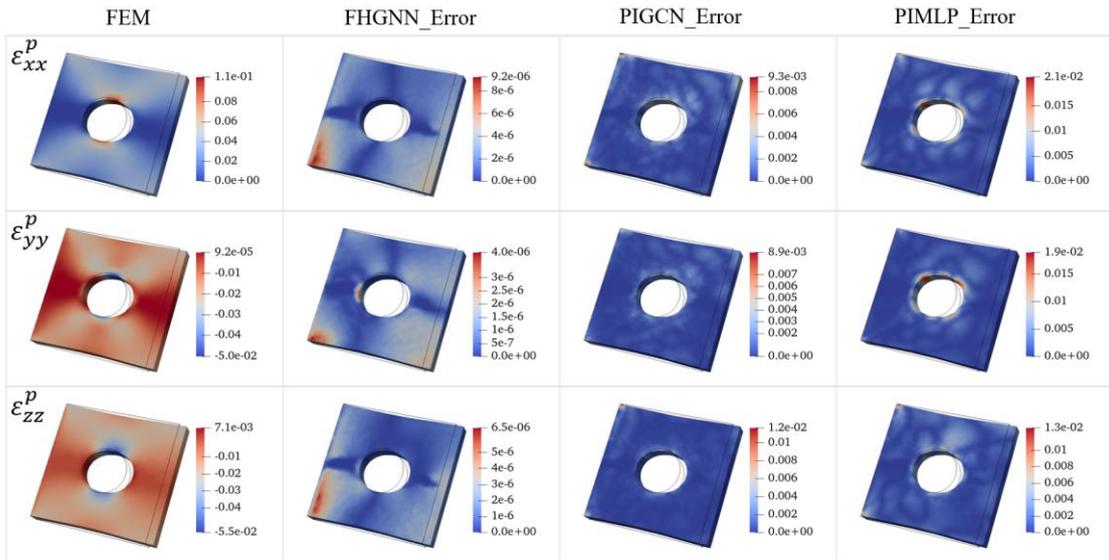

**Fig. 23.** 3D bi-material plate with a hole: Abaqus reference solution and pointwise absolute errors of FHGNN, PIGCN, and PIMLP for $\varepsilon_{xx}^p$, $\varepsilon_{yy}^p$ and $\varepsilon_{zz}^p$ (from top to bottom row).

**Table 6**
Performance comparison of PIMLP, PIGCN, and FHGNN on the 3D bi-material plate with a hole problem.

| Method | Time(s) | $L_2$ Error | | | | MAE | | | |
|---|---|---|---|---|---|---|---|---|---|
| | | $\sigma_{xx}$ | $\sigma_{yy}$ | $\sigma_{zz}$ | $\bar{\varepsilon}^p$ | $\sigma_{xx}$ | $\sigma_{yy}$ | $\sigma_{zz}$ | $\bar{\varepsilon}^p$ |
| PIMLP | 108.14 | 8.89E-02 | 7.07E-01 | 1.00E+00 | 8.77E-02 | 2.63E-01 | 3.10E-01 | 2.80E-01 | 1.47E-03 |
| PIGCN | 201.49 | 5.31E-02 | 3.05E-01 | 4.83E-01 | 3.38E-02 | 1.52E-01 | 1.51E-01 | 1.41E-01 | 6.61E-04 |
| FHGNN | 59.80 | 5.78E-05 | 1.77E-04 | 1.92E-04 | 6.31E-05 | 1.87E-04 | 9.12E-05 | 5.86E-05 | 1.38E-06 |

### 3.6 Further discussion

#### *3.6.1 Efficiency Comparison with FEM*

In addition to nonlinear materials, the high number of DOFs in practical engineering problems presents significant computational challenges. To assess computational performance, this study compares FEM and the proposed FHGNN across a range of DOFs. Using the linear isotropic hardening case in Section 3.3, five mesh densities were constructed, with training of the FHGNN terminating once the relative error fell below 1%. **Table 7** summarizes the computational times. While FEM benefits from parallel execution, the FHGNN leverages PyG's differentiable framework for automatic GPU parallelization. As shown in **Table 7**, FHGNN is less efficient than FEM for small-scale problems; however, as the degrees of freedom increase, its efficiency surpasses that of FEM running on multiple CPU cores, underscoring the method's potential for large-scale applications.

Similar to other fields of deep learning, transfer learning can be employed to further accelerate the proposed FHGNN. By leveraging related prior solutions, similar displacement fields can be used as initial guesses to enhance convergence. On the densest mesh configuration in **Table 7**, the FHGNN already outperforms FEM parallelized on multiple CPU cores. The results in **Fig. 24** illustrate that transfer learning can further amplify this advantage. Given the broad availability of related data, three cases are considered: (1) transfer from coarse-mesh solutions, (2) transfer across different materials, and (3) transfer across varying loading conditions. In all scenarios, transfer learning consistently accelerates convergence compared with training from scratch.

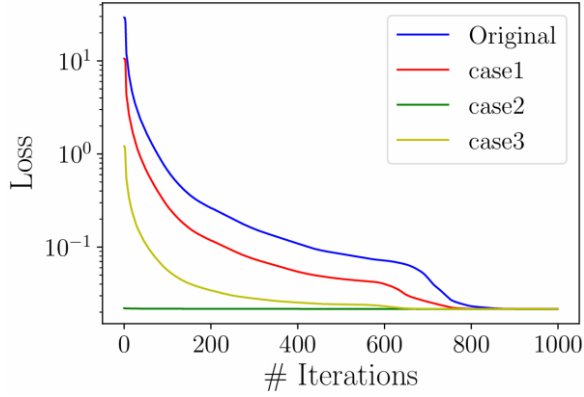

**Fig. 24.** The loss convergence histories after transfer learning from different sources.

**Table 7**
Computational efficiency across mesh resolutions for FEM (1-core CPU), FEM (8-core parallel), and the FHGNN.

| Mesh | FEM (1 core) | FEM (8 cores) | FHGNN |
|---|---|---|---|
| 40×10×10 | 2s | 1s | 1.936s |
| 80×20×20 | 47s | 8s | 12.18s |
| 120×30×30 | 157s | 45s | 47.39s |
| 160×40×40 | 1931s | 399s | 145.51s |
| 150×50×50 | / | 1021s | 286.92s |

### *3.6.2 Efficiency of FEM shape-function-based differentiation scheme*

Although AD is convenient and is natively supported by modern deep-learning frameworks, an increasing number of PINN-related studies have started to replace AD with numerical differentiation schemes, such as finite differences [44,45] or FEM shape functions [26]. Moreover, it was reported in [24] that FEM discretization and shape-function-based gradients can be more stable than AD and better mitigate strain-localization issues. Here, we further demonstrate that, shape-function-based differentiation also leads to improved training efficiency.

In PINNs, AD is employed to evaluate displacement gradients required by the governing equations. Here, we use a one-dimensional example and construct a neural network $\mathcal{N}^L(x)$ with a scalar input to approximate the displacement field. Let:

$$\mathcal{N}^0(x) = x, z^k(x) = W_k \mathcal{N}^{k-1}(x) + b_k, \tag{41}$$

$$\mathcal{N}^k(x) = \Phi\left(z^k(x)\right), 1 \leq k \leq L-1 \tag{42}$$

With the final layer:

$$\mathcal{N}^L(x) = W_L \mathcal{N}^{L-1}(x) + b_L \tag{43}$$

Where $z^k(x)$ denotes the pre-activation affine mapping. By the chain rule, the first derivative is:

$$\frac{\partial \mathcal{N}^L}{\partial x} = W_L \left( \prod_{k=L-1}^{1} \Phi'\left(z^k(x)\right) W_k \right) \quad (44)$$

In reverse-mode AD, the forward pass stores intermediate values $\{z^k(x), \mathcal{N}^k(x)\}$, and the backward pass computes vector–Jacobian products through this graph. When the loss include $\frac{d\mathcal{N}^L}{dx}$ (or higher spatial derivatives), AD must differentiate through the full derivative evaluation itself. This enlarges the computational graph and increases both memory usage and computational cost, since gradients must propagate through all operations involved in forming these spatial derivatives and their parameter dependencies $\theta = \{W_k, b_k\}_{k=1}^{L}$. As illustrated in **Fig. 25**, for the same scalar network $\mathcal{N}^L(x)$, the FD (first-order central-difference) and FEM-based differentiation schemes are respectively given by:

$$\left(\frac{d\mathcal{N}^L}{dx}\right)_i^{FD} \approx \frac{\mathcal{N}^L(x_{i+1}) - \mathcal{N}^L(x_{i-1})}{2h} \quad (45)$$

$$\left(\frac{d\mathcal{N}^L}{dx}\right)_j^{FEM} \approx \sum_{i=1}^{2} \left(\mathcal{N}^L(x_i) J^{-1} \frac{dN_i}{d\xi}\right) \quad (46)$$

For this one-dimensional problem, once the mesh and element type are fixed, $J$ and $\frac{dN_j}{d\xi}$ are precomputable constants. Consequently, in these two numerical-differentiation schemes, the derivative evaluation amounts to applying a fixed linear operator to the network outputs, so the resulting computational graph is essentially a linear extension of the original network graph, rather than the significantly more complex graph implied by Eq. (44). During backpropagation to update the network parameters, AD-based differentiation requires evaluating mixed higher-order derivatives such as $\partial_\theta (\partial_x \mathcal{N}^L(x))$, whereas numerical differentiation only requires the first-order parameter gradients $\partial_\theta \mathcal{N}^L(x)$. We argue that this structural simplification is a key reason why numerical differentiation is often more efficient in the PINN literature.

For the proposed FHGNN, the displacement field is no longer predicted by a neural network; instead, the nodal displacements $\boldsymbol{u}_j$ are treated as the unknown variables, rather than network parameters. The displacement gradient at a Gauss point within an element is given by:

$$\nabla_x u = \sum_j^{3} \left(\boldsymbol{u}_j ((J)^{-T} \nabla_\xi N_j)^T\right) \quad (47)$$

Once the mesh, element type, and Gauss points are fixed, $J$ and $\nabla_\xi N_j$ can still be treated as precomputable constants. Owing to the local-support property of FEM shape functions, evaluating the displacement gradient at a Gauss point can be viewed as a simple linear mapping, and the associated unknowns are sparse, involving only the nodal displacements of that element. In contrast, in PINNs the computational for displacement gradients typically depends on all unknown parameters (often exceeding 10E4, depending on the network architecture), whereas in FHGNN it involves only a small set of local variables (on the order of tens, depending on the element type). This locality is one of the key reasons for the high computational efficiency of the proposed method.

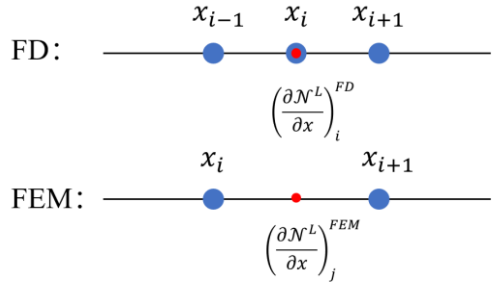

**Fig. 25.** Illustration of finite-difference and FEM-based differentiation schemes for evaluating spatial derivatives.

### *3.6.3 Analysis of convergence of different loss function*

In this section, we provide a detailed assessment of the energy-based and Galerkin-based loss functions. Their convergence behaviors are examined under controlled settings. Specifically, we consider the isotropic hardening cantilever beam in Section 3.3, the workpiece example in Section 3.4, and the bi-material plate with a hole in Section 3.5. Unless otherwise stated, all experimental configurations are kept identical, including the mesh resolution and optimizer. The FEM solution computed on the same mesh is used as the reference. We train FHGNN using the Galerkin loss with a sufficiently large number of training iterations to ensure convergence whenever possible, and the results are reported in **Table 8**. For the beam and workpiece cases, the Galerkin loss fails to converge to physically reasonable predictions even after extensive training. In contrast, for the plate-with-a-hole case, accurate predictions can be obtained after 36,090 training iterations. Notably, the energy-based loss achieves highly accurate predictions for all cases with fewer than 5,000 training iterations, demonstrating substantially faster and more reliable convergence.

We further investigate the influence of mesh resolution on the performance of different loss functions. Focusing on the isotropic hardening cantilever beam in Section 3.3, we conduct a set of controlled experiments with varying mesh densities. In all cases,

8-node linear hexahedral elements and structured meshes are used, with element counts of $20 \times 5 \times 5$, $40 \times 10 \times 10$, $80 \times 20 \times 20$, and $160 \times 40 \times 40$, corresponding to global mesh sizes of 0.2, 0.1, 0.05, and 0.025, respectively. Moreover, for all cases, the relative error is computed with respect to the FEM solution obtained on the densest mesh configuration of $160 \times 40 \times 40$, which serves as the reference. The results are summarized in **Fig. 26**, where the numbers annotated near the markers indicate the number of training iterations required for full convergence. We would expect the relative error of the Galerkin loss prediction to decrease under mesh refinement, although potentially at a slower rate than the energy loss. However, we observe the opposite trend: the prediction error under the Galerkin loss increases as the mesh is refined. This phenomenon is consistent with prior observations reported in [30]. From a numerical perspective, the discretization error should decrease with mesh refinement. We therefore attribute the deteriorating performance to the optimization/training behavior of the Galerkin loss, which appears to be sensitive to mesh density. In particular, denser meshes substantially increase the difficulty of optimization and lead to a sharp rise in the relative error. This mesh-dependent optimization degradation constitutes a major challenge for Galerkin loss and will be one of the focuses of our future work.

In above numerical testing cases, it is observed that, compared to the Galerkin loss, the energy loss exhibits superior convergence properties, along with higher efficiency and accuracy. To substantiate this conclusion, we provide a theoretical proof. In the proposed FHGNN, the nodal displacement vector $\boldsymbol{U}$ is treated as the unknown to be optimized within the network framework. Using an elastic constitutive model as an illustration, the discrete form of the energy loss can be rewritten as:

$$\mathcal{L}_{energy}(\boldsymbol{U}) = \frac{1}{2}\boldsymbol{U}^T\boldsymbol{K}\boldsymbol{U} - \boldsymbol{U}^T\boldsymbol{F}^{ext} \qquad (48)$$

The stiffness matrix $\boldsymbol{K}$ with Dirichlet boundary DOFs eliminated is real, symmetric, and positive definite [43]. With the eigenvalues of $0 < \lambda_{min} \leq \lambda_{max}$, the Euclidean($l_2$) condition number is given by:

$$\kappa(\boldsymbol{K}) = \frac{\lambda_{max}}{\lambda_{min}} \geq 1 \qquad (49)$$

In this framework, assuming gradient descent with a constant step size $\alpha > 0$, we have:

$$\boldsymbol{U}_{k+1} = \boldsymbol{U}_k - \alpha\nabla\mathcal{L}_{energy}(\boldsymbol{U}_k) = \boldsymbol{U}_k - \alpha\boldsymbol{K}(\boldsymbol{U}_k - \boldsymbol{U}^*) \qquad (50)$$

Given that $\boldsymbol{K}\boldsymbol{U}^* = \boldsymbol{F}^{ext}$, where $\boldsymbol{U}^*$ is the target solution. Denote the error as $\boldsymbol{e}^k = \boldsymbol{U}_k - \boldsymbol{U}^*$, we obtain the error update equation:

$$\boldsymbol{e}^{k+1} = \boldsymbol{U}_{k+1} - \boldsymbol{U}^* = (\boldsymbol{I} - \alpha\boldsymbol{K})\boldsymbol{e}^k \qquad (51)$$

By orthogonally diagonalizing $\boldsymbol{K} = \boldsymbol{Q}\boldsymbol{\Lambda}\boldsymbol{Q}^T (with\ \boldsymbol{\Lambda} = \text{diag}(\lambda_i))$, where $\lambda_i$ are the eigenvalues of $K$, and letting $y_k = Q^T e_k$. Multiplying by $Q^T$, we can rewrite the error update equation as:

$$y_{k+1} = (I - \alpha \Lambda) y_k \Rightarrow y_{k+1}^{(i)} = (1 - \alpha \lambda_i) y_k^{(i)} \tag{52}$$

Thus, the convergence rate is determined by the spectral radius:

$$\rho(I - \alpha K) = max_i |1 - \alpha \lambda_i| \tag{53}$$

Assume $\alpha = \frac{2}{\lambda_{max} + \lambda_{min}}$, which yields:

$$0 < \rho^* = \frac{\lambda_{max} - \lambda_{min}}{\lambda_{max} + \lambda_{min}} = \frac{\kappa(K) - 1}{\kappa(K) + 1} < 1 \tag{54}$$

After $k$ gradient updates, the error satisfies:

$$\|e_k\|_2 \leq \left(\frac{\kappa(K) - 1}{\kappa(K) + 1}\right)^k \|e_0\|_2 \tag{55}$$

Therefore, similar to numerical iterative algorithms, the convergence rate is influenced by the condition number; a larger condition number results in slower convergence. A very large condition number indicates that the linear system is ill-conditioned. In contrast, the Galerkin loss function can be written as (with the coefficient modified to $1/2$ for convenience in derivation):

$$\mathcal{L}_{galerkin}(U) = \frac{1}{2} \|KU - F^{ext}\|_2^2 \tag{56}$$

Substituting into Eq. (50), we obtain:

$$U_{k+1} = U_k - \alpha \nabla \mathcal{L}_{galerkin}(U_k) = U_k - \alpha K^T K (U_k - U^*) \tag{57}$$

The subsequent analysis follows the same reasoning as above, with the key difference being that for the Galerkin loss, we focus on another condition number $\kappa(K^T K)$. Based on $K = Q \Lambda Q^T$ and $Q^T Q = I$, the following expression can be derived:

$$K^T K = Q \Lambda \Lambda Q^T \tag{58}$$

Therefore, the eigenvalues of $K^T K$ are the squares of the eigenvalues of $K$, leading to:

$$\kappa(K^T K) = \left(\frac{\lambda_{max}}{\lambda_{min}}\right)^2 = (\kappa(K))^2 > \kappa(K) \tag{59}$$

When the condition number is amplified, the ill-conditioning becomes more pronounced, leading to slower convergence. Here, we demonstrate that the condition number of the Hessian matrix ($K^T K$) for the Galerkin loss function is larger than that of the Hessian matrix ($K$) for the Energy loss function, which explains why using the Energy functional as the loss function perform better.

In practical applications, we recommend using the energy loss function. However, it is important to note that the Galerkin loss function is more versatile, as deriving the energy functional for more complex plastic constitutive models can be difficult or even impossible. In this paper, we address the $J_2$ plasticity model, for which the energy functional can be easily derived. Preconditioning techniques aimed at reducing the condition number can be employed to improve the convergence of the Galerkin loss function, which we will investigate in future work. For interested readers, several recent studies that integrate preconditioning with PINNs may serve as useful references [46–48].

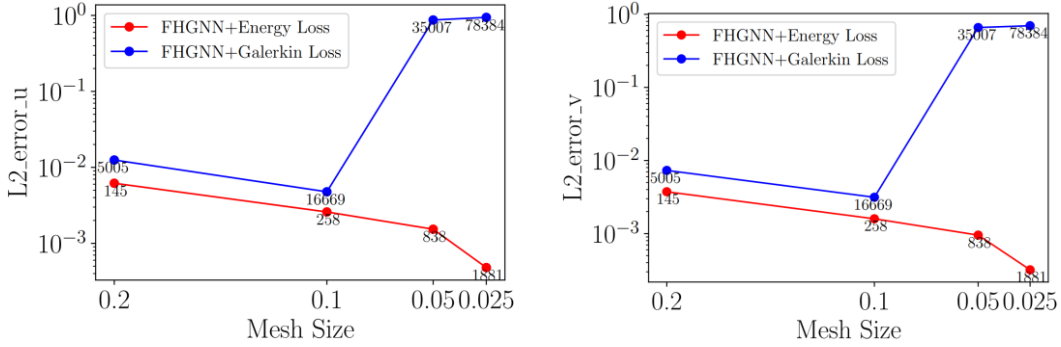

**Fig. 26.** The relative error after training to convergence for the two loss functions under different mesh resolutions.

**Table 8**

Comparison of training results for different loss functions across three cases.

| Case | Method | Epoch | Time (s) | $L_2\_u_x$ | $L_2\_u_y$ |
|---|---|---|---|---|---|
| Cantilever Beam | FHGNN+Energy loss | 5000 | 145.5 | 4.800E-04 | 3.194E-04 |
| | FHGNN+Galerkin loss | 78384 | 15753.9 | 9.417E-01 | 6.983E-01 |
| Workpiece | FHGNN+Energy loss | 5000 | 52.1 | 3.453E-03 | 5.708E-03 |
| | FHGNN+Galerkin loss | 525233 | 6825.1 | 1.483E+00 | 1.215E+00 |
| Plate with a hole | FHGNN+Energy loss | 5000 | 59.8 | 2.740E-05 | 1.371E-04 |
| | FHGNN+Galerkin loss | 36090 | 559.9 | 2.282E-04 | 8.200E-03 |

# 4 Conclusion and future work

We exploit the efficiency of GNNs and their natural compatibility with mesh data to embed an interpretable FEM computational pipeline into message passing. Specifically, we design three modules based on standard aggregation–update operations: (i) aggregating nodal geometry to construct element-wise geometric quantities, (ii) updating stresses and strains at element level, and (iii) assembling element contributions into nodal internal forces. Benefiting from FEM's discrete formulation and local-support property, displacement-gradients computation in FHGNN involves only sparse, local trainable variables, whereas generic AD tends to enlarge the computational graph and couple all network parameters. Motivated by empirical results and condition-number analysis, we adopt an efficient discrete variational loss. The resulting end-to-end differentiable formulation also supports optimizing arbitrary nodal attributes; a preliminary mesh-adaptivity study demonstrates this capability.

We benchmark FHGNN on multiple 3D problems, including footing, cantilever beam, workpiece, and a bi-material plate with a hole, covering cyclic loading with both isotropic and kinematic hardening. We compare against state-of-the-art PINN variants, including PIMLP and PIGCN (both are enhanced by FEM discretization). Across all cases, FHGNN predicts nonlinear elastoplastic responses with higher accuracy and efficiency, achieving on average over a $2\times$ speedup and improving accuracy by more than two orders of magnitude. With a GPU implementation, our method solves complex 3D plasticity problems within minutes and outperforms multi-core CPU FEM on dense meshes. These results position our method as a scalable alternative for path-dependent nonlinear systems.

Future work of the study will extend this framework beyond the current focus on closed-form $J_2$ plasticity. We plan to implement AD-compatible general return mapping algorithms for complex yield surfaces and explore multiphysics integration with DEM (for granular media) and FVM (for fluid dynamics) within the differentiable GNN framework. By embedding FEM computation into message passing, the proposed GNN acts as a data-free, physics-driven white-box solver: each step is physically interpretable, and its operations are prescribed rather than learned, which underpins the efficiency and robustness of the method. A key direction for future research is to learn interpretable yet more efficient message-passing modules [32] from large-scale data, potentially achieving accuracy–efficiency trade-offs beyond those attainable with standard FEM formulations. Further extensions will also evaluate operator learning synergies (e.g., FNO [49] and DeepONet [50]) for multiscale problems and investigate data assimilation techniques for experimental validation. Collectively, these directions advance our long-term vision of a unified framework bridging computational mechanics and deep learning for high-fidelity modeling of industrial-scale challenges.

## Data availability

All code and datasets used in this study will be made publicly available upon publication at https://github.com/yjc0416/FEM-Informed-Hypergraph-Neural-Networks.

## CRediT authorship contribution statement

**Jianchuan Yang:** Writing – original draft, review & editing, , Methodology, Formal analysis, Data curation, Visualization, Conceptualization. **Xi Chen:** Writing – review & editing, Conceptualization. **Jidong Zhao:** Writing – review & editing, Supervision, Funding acquisition.

## Conflict of interest

The authors declare that they have no known competing financial interests or personal relationships that could have appeared to influence the work reported in this paper.

## Acknowledgments

This work is financially supported by the National Natural Science Foundation of China (Key Project #52439001), and the Research Grants Council of Hong Kong (Nos. GRF 16203123, 16208224, 16214525,16217225, CRF C7085-24G, RIF R6008-24, TRS T22-607/24N, and T22-606/23-R).

# Appendix-A

## $J_2$ Plasticity

Considering the $J_2$ flow theory, the von Mises criterion is employed to describe yielding, incorporating both linear isotropic and kinematic hardening, as outlined below:

$$\begin{cases} f(\boldsymbol{\sigma}, \alpha, \boldsymbol{q}) = \|\boldsymbol{\eta}\| - \sqrt{\frac{2}{3}}(\sigma_Y + K\alpha) \\ \boldsymbol{\eta} = \boldsymbol{\sigma}' - \boldsymbol{q} \end{cases} \quad (1)$$

where $\boldsymbol{\eta}$ presents the relative stress, $\boldsymbol{\sigma}'$ denote the deviatoric parts of stress, and the back stress $\boldsymbol{q}$ defines the center of von Mises yield surface in stress deviator space. $\alpha$ is another internal plastic variable known as the equivalent plastic strain. Initial yield stress $\sigma_Y$ and the constant isotropic hardening modulus $K$ are used to compute the flow stress. The Karush–Kuhn–Tucker (KKT) conditions, consistency condition and evolution laws are defined as (with associative flow rule):

$$\begin{cases} \gamma \geq 0, \\ f(\boldsymbol{\sigma}, \alpha, \boldsymbol{q}) \leq 0, \\ \gamma f(\boldsymbol{\sigma}, \alpha, \boldsymbol{q}) = 0 \end{cases} \quad (2)$$

$$\gamma \dot{f}(\boldsymbol{\sigma}, \alpha, \boldsymbol{q}) = 0 \quad (3)$$

$$\begin{cases} \dot{\boldsymbol{\varepsilon}}^p = \gamma \dfrac{\boldsymbol{\eta}}{\|\boldsymbol{\eta}\|} \\ \dot{\alpha} = \sqrt{\dfrac{2}{3}}\gamma \\ \dot{\boldsymbol{q}} = \dfrac{2}{3}\gamma H \dfrac{\boldsymbol{\eta}}{\|\boldsymbol{\eta}\|} \end{cases} \quad (4)$$

where $\gamma$ denotes the consistency parameter and $H$ is the constant kinematic hardening modulus. $\dfrac{\boldsymbol{\eta}}{\|\boldsymbol{\eta}\|}$ gives the unit tensor normal to the yield surface.

To determine the stress that satisfies the nonlinear plastic constitutive relations, the radial return mapping algorithm [51] is employed. This method uses the stress and internal variables $[\boldsymbol{\sigma}_t, \boldsymbol{\varepsilon}_t^p, \alpha_t, \boldsymbol{q}_t]$ at step $t$, and the total $\boldsymbol{\varepsilon}_{t+1}$ or incremental strain $\Delta \boldsymbol{\varepsilon} = \boldsymbol{\varepsilon}_{t+1} - \boldsymbol{\varepsilon}_t$ to compute the state at next step. Initially, assuming elastic behavior, an elastic trial state is calculated:

$$\begin{cases} \boldsymbol{\sigma}'_{trial} = \boldsymbol{\sigma}'_t + 2\mu \Delta \boldsymbol{\varepsilon}' \\ \boldsymbol{\eta}_{trial} = \boldsymbol{\sigma}'_{trial} - \boldsymbol{q}_t \\ f_{trial} = \|\boldsymbol{\eta}_{trial}\| - \sqrt{\dfrac{2}{3}}(\sigma_Y + K\alpha_t) \end{cases} \quad (5)$$

where $\Delta\boldsymbol{\varepsilon}'$ is the deviatoric part of the strain increment. The yield condition is checked, if $f_{trial} \leq 0$, the assumption is acceptable. Otherwise, it indicates that the hypothetical state violates the yield criterion, necessitating further modifications:

$$\begin{cases} \Delta\gamma = \dfrac{f_{trial}}{2\left(\mu + \dfrac{H+K}{3}\right)} \\ \boldsymbol{n}_{t+1} = \dfrac{\boldsymbol{\eta}_{trial}}{\|\boldsymbol{\eta}_{trial}\|} \\ \boldsymbol{\varepsilon}^p_{t+1} = \boldsymbol{\varepsilon}^p_t + \Delta\gamma\boldsymbol{n}_{t+1} \\ \alpha_{t+1} = \alpha_t + \sqrt{\dfrac{2}{3}}\Delta\gamma \\ \boldsymbol{q}_{t+1} = \boldsymbol{q}_t + \dfrac{2}{3}\Delta\gamma H\boldsymbol{n}_{t+1} \\ \boldsymbol{\sigma}'_{t+1} = \boldsymbol{\sigma}'_{trial} - 2\mu\Delta\gamma\boldsymbol{n}_{t+1} \end{cases} \quad (6)$$

The stress calculation via the radial return mapping method involves conditional operations, with corrections applied only at certain gaussian points during yielding. Thanks to the design of AD, even conditional operations can be differentiated. The gradients are computed along different computational paths determined by the conditions, ensuring overall differentiability within the framework.